\documentclass[letteraper, 10 pt, journal, twoside]{support/IEEEtran}

\newcommand{\project}[0]{\pi}
\newcommand{\meanW}[0]{{^{W}\mathbf{p}}_{i}}
\newcommand{\meanC}[0]{{^{C}\mathbf{p}}_{i}}
\newcommand{\meanI}[0]{\mathbf{q}_{i}}
\newcommand{\meanJ}[0]{\mathbf{q}_{j}}

\newcommand{\covW}[0]{\boldsymbol{\Sigma}_{3D}}

\newcommand{\covI}[0]{\boldsymbol{\Sigma}_{2D}}

\newcommand{\nofG}[0]{^{W}\mathbf{n}_i}
\newcommand{\RofG}[0]{^{W}\mathbf{R}_i}

\newcommand{\TCW}[0]{{^{C}\mathbf{T}_{W}}}
\newcommand{\TWC}[0]{{^{W}\mathbf{T}_{C}}}

\newcommand{\TWI}[0]{{^{W}\mathbf{T}_{I}}}

\newcommand{\TLI}[0]{{^{L}\mathbf{T}_{I}}}

\newcommand{\TCI}[0]{{^{C}\mathbf{T}_{I}}}
\newcommand{\TIC}[0]{{^{I}\mathbf{T}_{C}}}

\newcommand{\transit}[0]{\mathbf{t}}

\newcommand{\tCW}[0]{{^{C}\transit_{W}}}

\newcommand{\tWI}[0]{^{W}\transit_{I}}

\newcommand{\tCI}[0]{^{C}\transit_{I}}

\newcommand{\RotCW}[0]{{^{C}\mathbf{R}_{W}}}
\newcommand{\RotWC}[0]{{^{W}\mathbf{R}_{C}}}

\newcommand{\RotWI}[0]{^{W}\mathbf{R}_{I}}

\newcommand{\RotIC}[0]{{^{I}\mathbf{R}_{C}}}
\newcommand{\RotCI}[0]{{^{C}\mathbf{R}_{I}}}

\newcommand{\Exp}[0]{\mathrm{Exp}}

\newcommand{\pd}[2]{\frac{\partial {#1} }{\partial {#2} }}

\newcommand{\SE}[1]{\boldsymbol{SE}(#1)}

\newcommand{\matJ}[0]{{\mathbf{J}_\pi}}

\newcommand{\matW}[0]{^{C}\mathbf{R}_W}

\newcommand{\rotlll}[0]{\delta \varphi}
\newcommand{\tlll}[0]{\delta t}

\newcommand{\rotrrr}[0]{\delta R}
\newcommand{\trrr}[0]{\delta \rho}

\IEEEoverridecommandlockouts

\usepackage[table]{xcolor}

\usepackage{makecell}
\usepackage{times}
\usepackage[pdftex]{graphicx}
\usepackage{subfigure}
\usepackage{amsmath,amssymb,amsopn,amstext,amsfonts}
\usepackage{cancel}
\usepackage[space]{cite}
\usepackage{siunitx}
\usepackage{soul}
\usepackage{caption}
\usepackage{lipsum,multicol}

\everymath{\displaystyle}
\usepackage{booktabs}
\usepackage{threeparttable}
\usepackage{amsthm}
\usepackage{xfrac}

\usepackage{mathtools}

\usepackage{tabularx}
\usepackage{bm}
\usepackage{diagbox}
\usepackage{float}
\usepackage{url}
\usepackage{multirow}
\usepackage{multicol}
\usepackage{tikz}
\usepackage{tablefootnote}

\usepackage[linkcolor=black,citecolor=black,urlcolor=black,colorlinks=true]{hyperref}
\usepackage{enumitem}
\usepackage[ruled,vlined,linesnumbered]{algorithm2e}
\usepackage{pifont}

\everymath{\displaystyle}

\DeclareMathAlphabet\mathbfcal{OMS}{cmsy}{b}{n}

\usepackage{cite}

\usepackage{soul}
\newcommand{\blind}[1]{\textcolor{black}{#1}}

\title{\LARGE \bf GS-LIVO: Real-Time LiDAR, Inertial, and Visual Multi-sensor Fused Odometry with Gaussian Mapping
}

\author{Sheng Hong$^{1,*}$,
Chunran Zheng$^{2,*}$,
Yishu Shen$^3$,
Changze Li$^3$,
Fu Zhang$^{2}$,
Tong Qin$^{3\dagger}$,
Shaojie Shen$^{1 }$ 
\thanks{
$*$ Equal contribution.
$^{\dag}$Corresponding author 
 (email: qintong@sjtu.edu.cn)}
\thanks{$^{1}$ Department of Electronic Computer Engineering, The Hong Kong University of Science and Technology, Hong Kong SAR, China. 
}
\thanks{$^{2}$ Department of Mechanical Engineering, The University of Hong Kong, Hong Kong SAR, China. 
}

\thanks{$^{3}$ Global Institute of Future Technology, Shanghai Jiao Tong University, Shanghai, China.
}
}

\begin{document}

\maketitle
\thispagestyle{empty}
\pagestyle{empty}

\begin{abstract}

% .

In recent years, 3D Gaussian splatting (3D-GS) has emerged as a novel scene representation approach. However, existing vision-only 3D-GS methods often rely on hand-crafted heuristics for point-cloud densification and face challenges in handling occlusions and high GPU memory and computation consumption~\cite{3dgstutorial}.

LiDAR-Inertial-Visual (LIV) sensor configuration has demonstrated superior performance in precise localization and dense mapping by leveraging complementary sensing characteristics: rich texture information from cameras, precise geometric measurements from LiDAR, and high-frequency motion data from IMU~\cite{xu2021fast,xu2022fast,lin2021r,lin2022r,r3live++,zheng2022fast,zheng2024fast}.

Inspired by this, we propose a novel real-time Gaussian-based simultaneous localization and mapping (SLAM) system. 
Our map system comprises a global Gaussian map and a sliding window of Gaussians, along with an IESKF-based real-time odometry utilizing Gaussian maps.
The structure of the global Gaussian map consists of hash-indexed voxels organized in a recursive octree. This hierarchical structure effectively covers sparse spatial volumes while adapting to different levels of detail and scales in the environment.
The Gaussian map is efficiently initialized through multi-sensor fusion and optimized with photometric gradients. 
Our system incrementally maintains a sliding window of Gaussians with minimal graphics memory usage, significantly reducing GPU computation and memory consumption by only optimizing the map within the sliding window, enabling real-time optimization.

Moreover, we implement a tightly coupled multi-sensor fusion odometry with an iterative error state Kalman filter (IESKF), which leverages real-time updating and rendering of the Gaussian map to achieve competitive localization accuracy.

Our system represents the first real-time Gaussian-based SLAM framework deployable on resource-constrained embedded systems (all implemented in C++/CUDA for efficiency), demonstrated on the \textbf{NVIDIA Jetson Orin NX}\footnote{Equipped with an 8-core CPU and 1024 CUDA cores, featuring 16\,GB LPDDR5 memory. For more details, see \url{https://www.nvidia.com/en-us/autonomous-machines/embedded-systems/jetson-orin-nx/}} platform. The framework achieves real-time performance while maintaining robust multi-sensor fusion capabilities.
All implementation algorithms, hardware designs, and CAD models and demo video of our GPU-accelerated system will be publicly available at 
\href{https://github.com/HKUST-Aerial-Robotics/GS-LIVO}{\textcolor{blue}{https://github.com/HKUST-Aerial-Robotics/GS-LIVO}}.

\end{abstract}

% \IEEEpeerreviewmaketitle
\section{Introduction}
% 一页
\begin{figure}[t]
    \centering
    \includegraphics[width=0.5\textwidth]{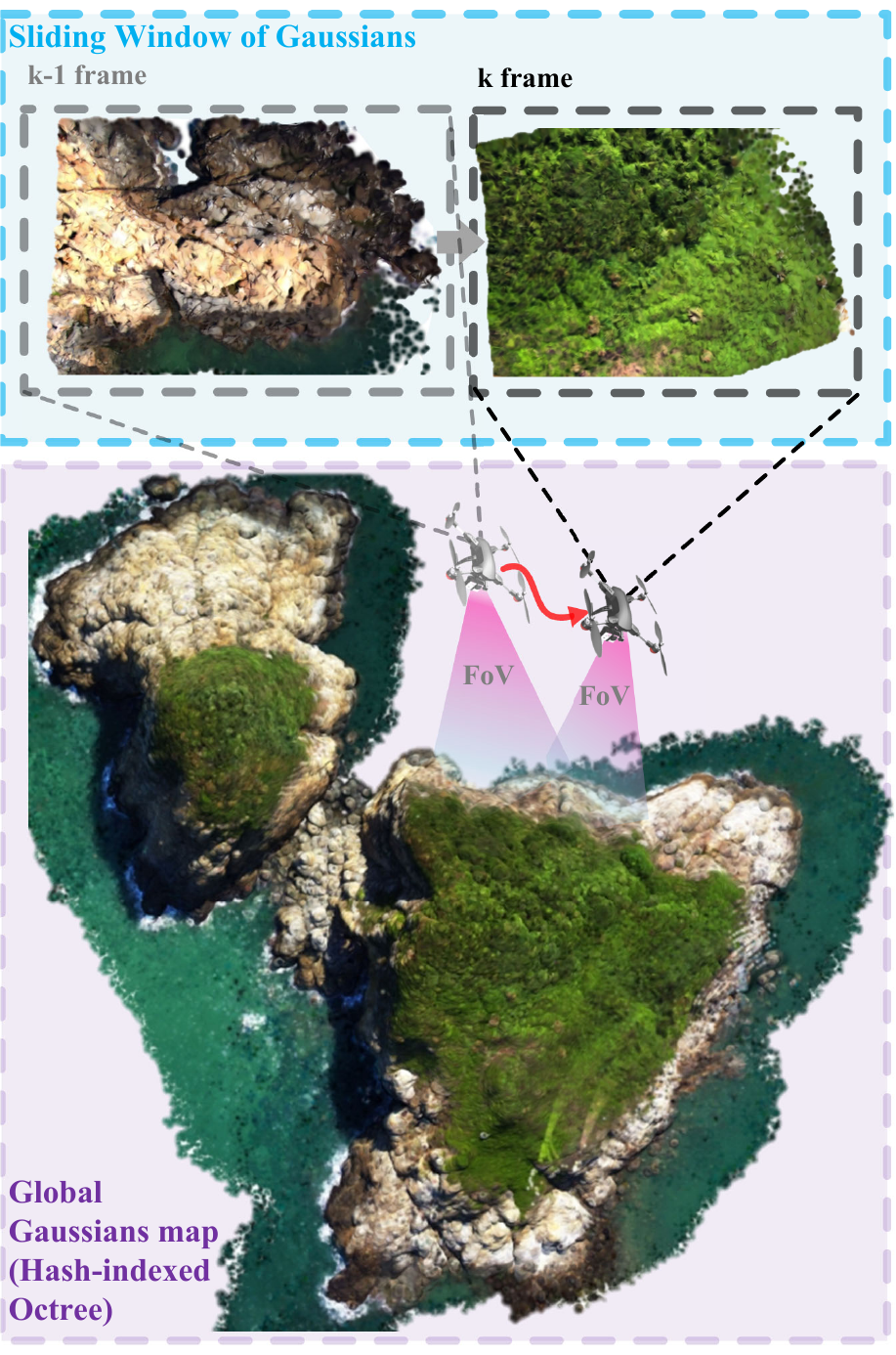}
    \caption{Components of GS-LIVO for Large-Scale Scenarios: Real-Time Odometry and Gaussian Mapping on Aerial Datasets~\cite{li2024mars}.}
    \label{fig:T1}
\end{figure}

\IEEEPARstart{I}{n} recent years, advances in Simultaneous Localization and Mapping (SLAM) have led to a variety of explicit map representations, including dense colored point clouds, sparse patch-based structures \cite{Forster17troSVO, zheng2022fast}, and even mesh-based \cite{lin2023immesh,jia2024cad,zhu2024mesh,ruan2023slamesh,wang2024simplified,wang2020pmds} or surfel-based \cite{cho2021sp,quenzel2021real,nguyen2023slict,wang2019real,gao2023surfelnerf} reconstructions. These forms, often integrated with feature-based or direct methods, support efficient, real-time operation across platforms such as UAVs and mobile robots \cite{gao2020autonomous, qin2018vins, xu2021fast, xu2022fast, lin2021r, lin2022r, zheng2022fast}. Many state-of-the-art SLAM systems leverage these classical map constructs due to their well-established pipelines and robust performance in pose estimation tasks.
However, while such handcrafted, explicit representations have matured substantially, certain limitations remain. They typically rely on abundant geometric features and high-frame-rate inputs to ensure stable tracking. Moreover, these methods often struggle to provide photorealistic reconstructions and are generally confined to explaining only the observed parts of a scene. This shortfall poses challenges in applications that require the prediction or synthesis of new viewpoints, such as immersive augmented reality, high-quality 3D modeling, and scenarios where unseen regions must be inferred for robust decision-making.
Recently, breakthroughs in novel view synthesis have introduced neural representations capable of photorealistic rendering from arbitrary viewpoints. Implicit models like Neural Radiance Fields (NeRF) \cite{mildenhall2021nerf} and explicit structures such as 3D Gaussian Splatting (3DGS) \cite{kerbl20233d, hierarchicalgaussians24} not only enrich the fidelity of the reconstructed environment but also open the door to more advanced SLAM paradigms. A wave of NeRF-based SLAM approaches \cite{SucaretalICCV2021,Zhu_2022_CVPR,Wang_2023_CVPR,Johari_2023_CVPR,Sandstrom2023UncLeSLAM,Chung2022OrbeezSLAM} and 3DGS-based methods \cite{keetha2024splatam,yan2023gs,sun2024mm3dgs,xiao2024liv,hong2024liv} seek to integrate these high-fidelity representations into the SLAM pipeline, aiming to exploit the richer photometric and geometric cues for more accurate localization and mapping. Intuitively, as higher-quality maps provide better spatial and appearance cues, they should enhance the accuracy and robustness of pose estimation, thereby reinforcing the reciprocal relationship between mapping and localization within SLAM.
Despite their promise, current neural-based SLAM systems face a key bottleneck: maintaining truly real-time map updates. Although some approaches achieve near real-time odometry, their map optimization and refinement processes often lag behind, relying on separate, slower threads \cite{Zhu_2022_CVPR, keetha2024splatam, sun2024mm3dgs}. This mismatch between fast pose estimation and slower map updating reduces the system’s adaptability, particularly in dynamic or large-scale environments where continuously refreshed, high-fidelity maps are essential. As a result, the practical deployment of these high-quality representations in robotics—where rapid environmental interpretation is paramount—remains limited.

Motivated by these issues, this work aims to develop a LiDAR-Inertial-Visual Odometry (LIVO) system that tightly integrates LiDAR and camera measurements using a novel map representation of Gaussian. The objectives of this work are twofold: achieving high-precision localization through tightly coupled multisensor fusion within a Gaussian map, and significantly improving the efficiency of Gaussian map updates, even in large-scale (as exemplified by the aerial scene shown in Fig.~\ref{fig:T1})  and complex environments. This approach addresses key bottlenecks that hinder the practical deployment of current Gaussian-SLAM systems.

In summary, our key contributions include:

\begin{itemize}
\item{We introduce a global Gaussian map representation structured as a spatial hash-indexed octree. This hierarchy structure enables efficient global indexing, co-visibility checks, and inherently supports varying Levels of Detail (LoD) to accommodate large-scale scenes (Section~\ref{sec.over}).}
\item{We propose a fast initialization strategy that fuses LiDAR and visual data to rapidly generate a well-structured global Gaussian map with high-fidelity rendering (Section~\ref{sec.init})}
\item{We propose an sliding window of Gaussians to incremental method scheduling the Gaussians, which can minimize the cost in maintaining the map and compuation burden optimize of Gaussians and reduce GPU memory burden (Section~\ref{sec.map})}
\item{We propose a novel visual measurement model leveraging the photorealistic rendering capabilities of our Gaussian Map, tightly integrating LiDAR-inertial measurements using an IESKF with sequential updates (Section~\ref{sec.state}).}
\end{itemize}

Extensive benchmark and real-world experiments (Section~\ref{sec.Experiment}) show that our method significantly reduces memory usage and accelerates Gaussian map optimization, while achieving competitive odometry accuracy and maintaining high rendering quality across both indoor and outdoor datasets.

\subsection{Neural Radiance Fields (NeRF) based SLAM}

Recent advancements in SLAM technology have seen significant developments in the use of Neural Radiance Fields (NeRFs) for implicit map representations. iMAP \cite{SucaretalICCV2021} pioneered the application of implicit maps in SLAM systems, marking a notable milestone despite not outperforming traditional visual odometry methods. Building on iMAP's foundational ideas, NICE-SLAM \cite{Zhu_2022_CVPR} introduced an open-source solution with enhanced scalability. It employs multi-scale multilayer perceptrons (MLPs) to model geometric structures at various scales, utilizing error-guided probability for pixel-based frame sampling, which significantly improves efficiency. Furthermore, Instant-NGP~\cite{mueller2022instant} addresses the computational intensity of NeRF networks by introducing innovative position encoding, reducing network size, and enhancing overall performance through effective multi-resolution hashing.

Co-SLAM \cite{Wang_2023_CVPR} leverages map representations to learn spatial geometric structures across different frequencies, thereby enhancing both mapping and localization precision. ESLAM \cite{Johari_2023_CVPR}, similar to NICE-SLAM, focuses on a prior-free training model and uses signed distance function (SDF) maps and carefully designed loss functions to reduce memory consumption from cubic to quadratic growth using axis-aligned planes. UncLeSLAM \cite{Sandstrom2023UncLeSLAM} builds upon NICE-SLAM by modeling pixel uncertainty through Laplacian uncertainty, which, although slightly increasing computational time and memory usage, significantly improves accuracy via self-supervised Laplacian uncertainty modeling.

Lastly, Orbeez-SLAM \cite{Chung2022OrbeezSLAM} integrates the traditional ORB-SLAM\cite{ORBSLAM3_TRO} with a NeRF map from Instant-NGP, demonstrating the successful fusion of classical and modern implicit map representations. More recent works, such as H$_2$-Mapping and H$_3$-Mapping \cite{h2mapping2023,h3mapping2023}, further push the limits of real-time capability and high-quality reconstruction on edge devices by introducing hierarchical hybrid representations, improved initialization schemes, and advanced keyframe selection strategies. Similarly, Swift-Mapping \cite{swiftmapping2023} employs a neural implicit octomap structure to achieve efficient neural representation of large, dynamic urban scenes, enabling online updates and significantly accelerating reconstruction speeds.

These contributions collectively showcase the potential of NeRF-based SLAM in enhancing the robustness, efficiency, and accuracy of 3D scene reconstruction and localization. However, while NeRF-based methods effectively reduce memory consumption through implicit representations, they often require substantial optimization time, making it challenging to achieve high-fidelity, high-frame-rate mapping updates.

\subsection{3D Gaussian Splatting in SLAM}
3D Gaussian Splatting (3DGS)\cite{kerbl20233d} is an explicit representation method for scene modeling and rendering, which utilizes 3D Gaussian to depict the geometric structure and appearance of a scene. This approach excels in novel view synthesis and real-time rendering, significantly reducing parameter complexity compared to traditional representations such as meshes or voxels. To further enhance the efficiency and scalability of 3D Gaussian representations, some studies have focused on compression techniques. For instance, Motion-GS\cite{guo2024motiongs} and RTG-SLAM\cite{peng2024rtgslam} achieve large-scale 3D reconstruction by employing compressed Gaussian maps, providing effective representations more suitable for real-time applications. Level of Detail (LoD) techniques are also utilized to manage the complexity of rendering large 3D scenes. Octree-GS\cite{ren2024octree} organizes data into a hierarchical structure, supporting multi-resolution anchors that ensure stable rendering speeds from different viewpoints. The Level of Gaussians (LoG)\cite{shuai2024LoG} employs a tree-based hierarchical structure, dynamically selecting appropriate detail levels based on the observer's distance and viewpoint, effectively allocating resources while preserving details in close-up views. These LoD techniques efficiently address computational bottlenecks encountered when rendering extensive and complex 3D scenes, making them highly suitable for real-time and large-scale applications.

Moreover, to improve the initialization results of 3D Generalized State (3DGS) systems, several works based on LiDAR point clouds have been proposed. 
LIV-GaussMap~\cite{hong2024liv} and Gaussian-LIC~\cite{lang2024gaussian} are among the earliest methods that utilize LiDAR for initializing the structure of maps, providing prior pose estimation, and further optimizing the map using photometric gradients.
LiV-GS\cite{xiao2024liv}further incorporates constraints such as normal vector loss to optimize the map.
LetsGo\cite{LetsGo} introduces a handheld polar coordinate scanner for capturing RGBD data of large parking environments, combined with LiDAR-assisted Gaussian primitives, achieving high-quality large-scale garage modeling and rendering. LI-GS\cite{jiang2024li}, on the other hand, enhances geometric accuracy in large-scale scenes by converting LiDAR data into plane-constrained multimodal Gaussian Mixture Models (GMMs). This method uses GMMs during both the initialization and optimization stages to ensure sufficient and continuous supervision over the entire scene, thereby mitigating the risk of overfitting.

In recent years, 3D Gaussian Splatting (3D-GS) has found widespread application as a mapping representation method in the field of Simultaneous Localization and Mapping (SLAM). 3D-GS has demonstrated significant improvements in real-time performance, map rendering, iterative updates, and partial support for dense RGB-D SLAM. For example, PhotoSLAM\cite{hhuang2024photoslam} uniquely combines explicit geometric and photometric features, integrating the characteristics of ORB-SLAM\cite{ORBSLAM3_TRO} with traditional techniques such as Gaussian pyramid representation and superpixels, emphasizing photorealistic map construction. Mono-GS\cite{MonoGS} integrates 3D Gaussian representation with a real-time differentiable splatting pipeline, introducing depth loss and manual pose Jacobian to improve pose estimation accuracy. GS-SLAM\cite{yan2023gs} leverages 3D Gaussian representation to balance efficiency and accuracy, achieving scene reconstruction and robust pose tracking through adaptive expansion strategies, with a notable speedup compared to neural implicit methods. SplaTAM\cite{keetha2024splatam} adopts a contour-guided optimization approach, dynamically expanding the map capacity based on rendered contours and input depth, ensuring efficient and accurate map updates in dynamic environments. MM3DGS SLAM\cite{sun2024mm3dgs} proposes a multi-modal SLAM framework that utilizes visual, inertial, and depth measurement data, achieving efficient scene reconstruction and real-time rendering through 3D Gaussian representation.

However, most existing 3D-GS-based SLAM solutions focus on real-time odometry rather than maintaining high-frequency, real-time map updates, causing map construction to lag behind sensor acquisition. Such low update rates critically limit a robot’s ability to swiftly interpret its surroundings and navigate dynamic or large-scale environments. In contrast, our approach achieves consistently high-frequency map updates—over 10 Hz indoors on a single CPU thread and around 3 Hz outdoors—thereby setting a new benchmark for 3D-GS-based SLAM and significantly enhancing its practical utility in real-world robotic applications.

\begin{figure*}[ht]
    \centering
    \includegraphics[width=1.0\textwidth]{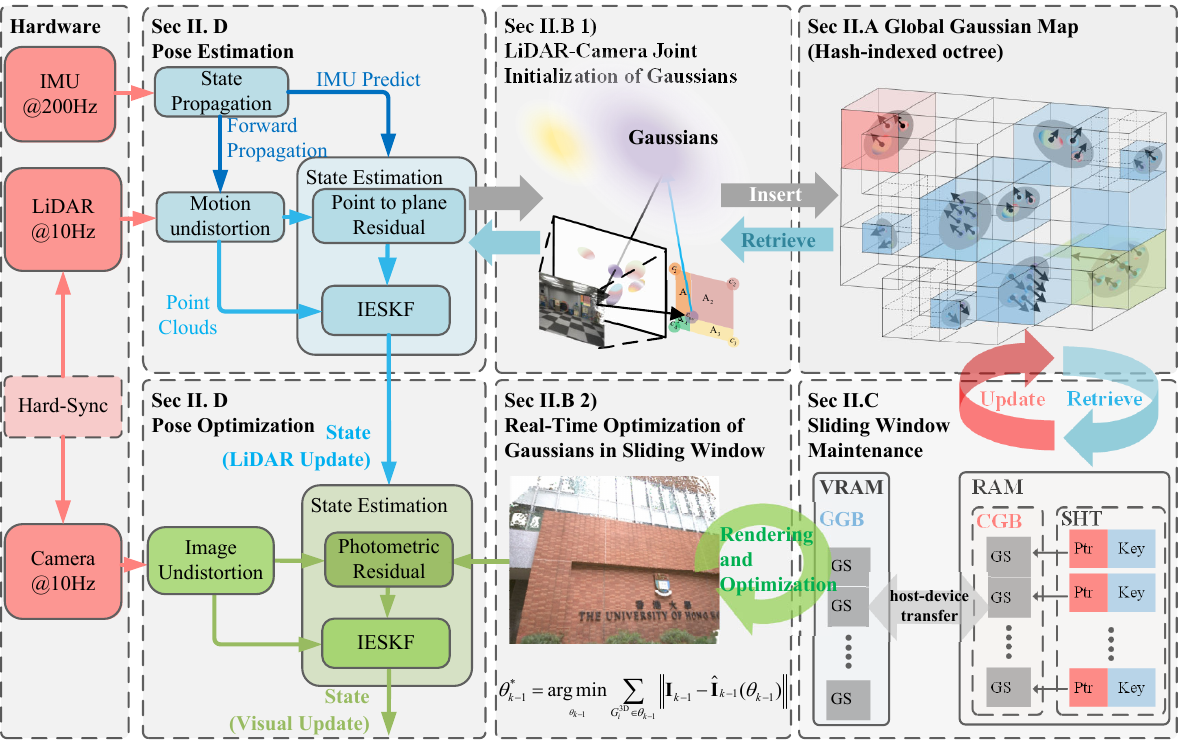}
    \caption{System overview of GS-LIVO: a real-time LiDAR-Inertial-Visual odometry system with Gaussian Splatting-based mapping. The pipeline performs joint initialization and optimization of Gaussians using multi-sensor data, managed through a hash-indexed octree structure and sliding window mechanism.}
    \label{fig:Overview}
\end{figure*}

\subsection{LiDAR-Inertial-Visual Multi-Sensor Fusion in SLAM}

In the field of Simultaneous Localization and Mapping (SLAM), recent advancements have significantly propelled the development of multi-sensor fusion, particularly in the integration of Light Detection and Ranging (LiDAR), Inertial Measurement Units (IMUs), and visual sensors. One pioneering work in this domain is LIC-Fusion\cite{zuo2019lic}, which effectively combines LiDAR, IMU, and visual data to enhance the overall performance of SLAM systems. By leveraging the complementary characteristics of these sensor modalities, LIC-Fusion establishes a robust foundation for subsequent research.

Building upon this groundwork, CamVox\cite{zhu2021camvox} further refines the integration process, providing a more resilient and efficient solution for real-time SLAM. This system capitalizes on the unique strengths of each sensor type, achieving superior accuracy and reliability through an optimized fusion strategy that minimizes redundancy while maximizing information gain.

Subsequently, LVI-SAM\cite{shan2021lvi} introduces a tightly coupled approach that integrates a Visual-Inertial System (VIS) with a LiDAR-Inertial System (LIS). Utilizing factor graphs and sliding window optimization, LVI-SAM significantly enhances the robustness and precision of the SLAM system, making it well-suited for challenging environments where high accuracy and reliability are critical. The framework's ability to handle large-scale mapping tasks under dynamic conditions highlights its versatility and effectiveness.

R$^2$LIVE\cite{lin2021r} advances the state-of-the-art by employing an error-state iterated Kalman filter to reduce feature reprojection errors. This technique ensures real-time performance and robustness, even in highly dynamic and complex scenarios. By addressing the challenges associated with sensor synchronization and drift, R$^2$LIVE demonstrates improved consistency and stability in the estimated trajectories.

R$^3$LIVE\cite{lin2022r} represents another significant step forward, introducing a novel method based on photometric error for constructing dense point cloud maps. Unlike traditional feature-based approaches that heavily rely on prominent visual features, R$^3$LIVE reduces computational cost by selecting points with larger gradients. This method reduces feature extraction computations but is prone to local optima due to point-based photometric errors and requires maintaining a dense colored global map continuously.

FAST-LIVO\cite{zheng2022fast, zheng2024fast}, compared with the colored point-based map maintained by R$^3$LIVE, employs a unified surfel-based map composed of adaptive-sized planes that integrate both LiDAR and visual map points. Sparse visual points are attached with image patches, enabling patch-based photometric errors with superior convergence properties. Inspired by its map structure, we design our own global Gaussian map. However, our Gaussian map features a higher density of map points to support dense view synthesis. To maintain efficiency and prevent excessive memory usage, we adopt Level of Detail (LoD) techniques and a Gaussian map sliding mechanism.

\begin{table}[t]
    \centering
    \caption{Notation and Symbol Definitions}
    \label{tab:symbols_all}
    \begin{tabular}{ll}
    \toprule
    Notation & Explanation \\
    \midrule
    %% 坐标与变换
    ${^W(\cdot)}$ & Vector expressed in the world frame \\
    ${^C(\cdot)}$ & Vector expressed in the camera frame \\
    ${^I(\cdot)}$ & Vector expressed in the IMU frame \\
    ${^L(\cdot)}$ & Vector expressed in the LiDAR frame \\
    $\SE{3}$ & Special Euclidean group in 3D spaces \\
    $\TLI \in \SE{3}$ & LiDAR-to-IMU extrinsic \\
    $\TIC \in \SE{3}$ & IMU-to-camera extrinsic  \\
    $\TWI \in \SE{3}$ & IMU pose w.r.t. world  \\
    $\TWC \in \SE{3}$ & Camera pose w.r.t. world  \\
    %% Gaussian
    $\mathcal{N}(\meanW,\covW)$ & 3D Gaussian defined in world coordinates \\
    $\mathcal{N}(\meanI,\covI)$ & 2D Gaussian after projection to image plane \\
    $\meanW,\meanI$ & Gaussian mean in 3D / 2D  \\
    $\covW,\covI$ & Gaussian covariance in 3D / 2D \\
    $\RofG$ & Gaussian orientation (rotation) matrix in 3D \\
    $\nofG$ & Normal vector of the objects surface  \\
    $\mathbf{S}(\cdot)$ & Scaling matrix indicating the spatial extent along axes \\
    $\sigma_i$ & Opacity of the $i$-th Gaussian \\
    $\widehat{\mathbf{I}}(\cdot), \mathbf{I}(\cdot)$ & Rendered image / Captured image \\
    %% 地图结构
    $\mathbf{v_s}$ & Root voxel length in the spatial structure \\
    %% 优化与估计
    $\project(\cdot)$ & Projection model of the pinhole camera\\
    $\matJ$ & Jacobian matrix of the projection model \\
    $\TWC^{*}$ & Optimized camera pose after optimization \\
    $\theta_{k-1},\theta_{k-1}^{*}$ & Gaussian parameters before/after optimization \\
    \bottomrule
    \end{tabular}
\end{table}

\section{Methodology\label{sec.Methodology}}
The system overview of GS-LIVO is illustrated in Fig. \ref{fig:Overview}. The hardware configuration integrates synchronized LiDAR, IMU, and camera, with precise temporal alignment ensured via an emulated pulse-per-second (PPS) signal~\cite{hong2023rollvox, zheng2022fast, zheng2024fast}. The software framework comprises four key modules: 
(1) a global Gaussian map organized with a spatial hash-indexed octree that effectively covers sparse spatial volumes while adapting to various environmental details and scales (Section~\ref{sec.Methodology}.A); 
(2) rapid initialization and online optimization of Gaussians based on LiDAR and visual information with photometric gradients (Section~\ref{sec.Methodology}.B); 
(3) incremental maintenance of sliding windows of local Gaussians for real-time optimization with minimal graphical memory usage (Section~\ref{sec.Methodology}.C); and (4) pose optimization using an IESKF with sequential updates (Section~\ref{sec.Methodology}.D). Our system represents a real-time SLAM framework that seamlessly integrates LiDAR, inertial, and visual sensors to achieve competitive localization accuracy. For clarity of presentation, we first define the notations used throughout this section in Table~\ref{tab:symbols_all}, which summarizes the coordinate transformations, Gaussian attributes, spatial structures, and optimization parameters.

\subsection{The Global Gaussian Map: Hash-indexed Octree\label{sec.over}}

% Gaussian local sliding windows 和local Gaussian  sliding windows 用哪个
Fig.~\ref{fig:iGM} illustrates the structure of the global map and the sliding strategy for Gaussian within our system. The mapping system consists of two main components: the global Gaussian map and the sliding windows of Gaussians. The global Gaussian map utilizes a hash-indexed octree structure that employs spatial hashing to efficiently cover the sparse volumes of the scene. This structure can recursively subdivide regions based on environmental complexity, enabling a more detailed and fine-grained map representation. The indexing of root voxels is determined based on spatial hash keys, calculated as follows:

\begin{equation}
\text{HashKey} = \left\lfloor \frac{\meanW}{\mathbf{v_s}} \right\rfloor 
\end{equation}

where $\mathbf{v_s}$ is the length of the root voxel. This approach allows for efficient management and indexing of sparsely distributed data points in large-scale environments.

In our LiDAR-Inertial-Visual system, we use the LiDAR data from the current frame to compute the spatial key, thereby efficiently identifying the root voxels in the global Gaussian map that correspond to the current field-of-view (FoV). This method facilitates the identification of co-visibility associations within the spatial environment, determining which areas are observable from the current viewpoint, thus providing accurate map information for subsequent processing. However, due to the non-contiguous nature of hash-indexed storage, direct GPU processing of Gaussian parameters for parallel optimization is constrained. A non-contiguous memory layout leads to inefficiencies in GPU access and processing, introducing unnecessary latency.

To address this issue, we have designed a specialized sliding window of Gaussian specifically for maintaining the Gaussian within the FoV, with a contiguous memory layout to ensure efficient GPU processing of Gaussian parameters. Specifically, the entire Gaussian of the large-scale environment is stored in random access memory (RAM) using a non-contiguous hash-octree structure in the global Gaussian map, while only the Gaussian located within the FoV are stored in a contiguous RAM region, with a duplicate kept in video random access memory (VRAM). When optimization of Gaussian parameters is required, these parameters are transferred from RAM to VRAM, allowing the GPU to perform parallel optimization efficiently. Upon completion of the optimization process, the updated Gaussian is transferred back from VRAM to RAM to maintain consistency across the global map.

Unlike the typically limited and non-scalable graphics memory, RAM provides greater capacity and can be easily expanded via swap space, enabling the handling of larger and more complex scenes. This design not only enhances the efficiency of GPU processing but also ensures the stability and reliability of the system when dealing with large-scale datasets, as shown in Fig.~\ref{fig:iGM}.

\subsection{Initialization and Optimization of Gaussians\label{sec.init}}

When a new LiDAR and camera frame is received, voxel-based map down-sampling on dense LiDAR point is implemented to mitigate GPU memory consumption.
Unlike others such as~\cite{keetha2024splatam,yan2023gs,MonoGS} that use boundaries for selection, we employ the leaf nodes voxels of the octree, to sample the object's surface in 3D space. By selecting LiDAR points within each voxel, the scene is represented more efficiently.

If the leaf voxel is not filled to capacity, new Gaussians will be roughly initialized with LiDAR and camera, as shown in Fig. \ref{fig:Overview}, and inserted into the leaf voxel.

\subsubsection{LiDAR-Camera Joint Initialization of Gaussians\label{sec.inita}}

In this step,the structural parameters are initialized with LiDAR.
Specifically, the scaling matrix $\mathbf{S}$ for the Gaussians is initialized based on the level of the hash-voxel, expressed as follows:
\begin{equation}
\mathbf{S}_i(\mathbf{s})
= \begin{pmatrix}
\text{s}_\delta & 0 & 0 \\
0 & \text{s}_y& 0 \\
0 & 0 & \text{s}_z
\end{pmatrix}
    \label{eq:scaling_matrix}
\end{equation}
Notably, the Gaussian is modeled as a 2D planar attached to surfaces of objects.
The $\text{s}_\delta$ is a hyper-parameter representing a minor numerical value to establish the slice feature.

The rotation matrix of Gaussian is initialized using the normal vector $\nofG$ of the surface, which is derived from the LiDAR-inertial SLAM system~\cite{yuan2022efficient}.
\begin{equation}
\RofG = \begin{pmatrix} 
\frac{\mathbf{e}_x\times \nofG}{\|\mathbf{e}_x\times \nofG\|} & 
\nofG \times \left(\frac{\mathbf{e}_x \times \nofG}{\|\mathbf{e}_x\times \nofG\|}\right) & 
\nofG
\end{pmatrix}
\end{equation}
where $\mathbf{e}_x $ is the normal vector on x-axis.

Lastly, the covariance matrix of Gaussian is constructed as follows:

\begin{equation}
\covW{}_i = (\RofG\mathbf{S}_i)(\RofG\mathbf{S}_i)^{T}
\label{equ:3dgs}
\end{equation}
During the rasterization procedure, the influence of $\alpha_i$ is determined using the product of the 2D Gaussian $\mathcal{N}(\meanI, \covI)$ and opacity $\sigma_i $. The formation of the 2D Gaussian involves splatting a 3D Gaussian $\mathcal{N}(\meanW, \covW)$ onto the screen space, as illustrated in Equation \eqref{equ:q} and \eqref{equ:spat}.

\begin{equation}
{\meanI}=\project(\TCW\meanW)
\label{equ:q}
\end{equation}
\begin{equation}
\covI{}_i = (\matJ \RotCW) \covW{}_i (\matJ\RotCW )^{T}
\label{equ:spat}
\end{equation}
where, $\project$ denotes the projection transformation. The linear approximation of the projective transformation $\project$ is denoted by the Jacobian $\matJ$.
In addition, $\TCW \in \SE{3}$ represents the transformation from the world frame to the camera frame, with $\RotCW$ denoting its rotational component.

\subsubsection{Real-Time Optimization of Gaussians in Sliding Window\label{sec.opt}}

The zero order of the spherical harmonic coefficients is initialized with bilinear interpolation on the image captured by the camera, While the higher order is initialized with zero.

As illustrated by the equation below, bilinear interpolation is utilized to compute the color of projected non-integral pixels.
\begin{equation}
\mathbf{c}(\meanI) = \sum_{j=1}^{4} \mathbf{c}_j \cdot A_j
\end{equation}
The computed value in \( c(\meanI) \) represents a weighted sum of the colors of the adjoining pixels. The weighting factor \( A_i \) is associated with the area corresponding to its neighboring pixels.

After initialization, the Gaussian will be further optimized with photometric gradients.
The optimization process is outlined as follows:

Firstly, the Gaussians render an image $\widehat{\mathbf{I}}_{k}$ with alpha blending, in accordance with:

\begin{equation}
\widehat{\mathbf{I}}_{k}(\meanI) = \sum_{i=1}^{M}
 [c_{i} {\sigma_i} G^\text{2D}_i(\meanI) \prod_{j=1}^{i-1}(1-\sigma_j G^\text{2D}_j(\meanJ))
 ]
\label{eq:raster}
\end{equation}
where, $G^\text{2D}_i(\meanI,\covI)$ signifies the two-dimensional Gaussian derived from $G^\text{3D}_i(\meanC,\covW)$ by applying a pose transform and local affine transformation~\cite{zwicker2001surface} illustrated in Equation \eqref{equ:spat}.
The parameter $\sigma_i \in [0,1]$ represents the opacity related to the Gaussians, and $M$ indicates the number of Gaussians influencing the pixel.

To refine the Gaussian map parameters, we consider the following optimization problem:
\begin{equation}
\theta_{k-1}^{*} = \arg\min_{\theta_{k-1}} \sum_{G^\text{3D}_i\in \theta_{k-1}} \Big\| \mathbf{I}_{k-1}- \widehat{\mathbf{I}}_{k-1}(\TWC; \theta_{k-1}) \Big\|
\label{eq:gaussian_optimization}
\end{equation}
where \(\theta_{k-1}\) denotes the structure parameters and spherical harmonic coefficients of the Gaussian elements within the current field of view. By minimizing this photometric loss, we iteratively adjust \(\theta_{k-1}\) such that the rendered image \(\widehat{\mathbf{I}}_{k-1}\) better matches the observed image \(\mathbf{I}_{k-1}\). We employ the Adam optimizer to efficiently solve this problem, updating the Gaussian parameters to achieve a more accurate and visually consistent map representation.

\begin{figure*}[t!]
    \centering
    \includegraphics[width=1.0\textwidth]{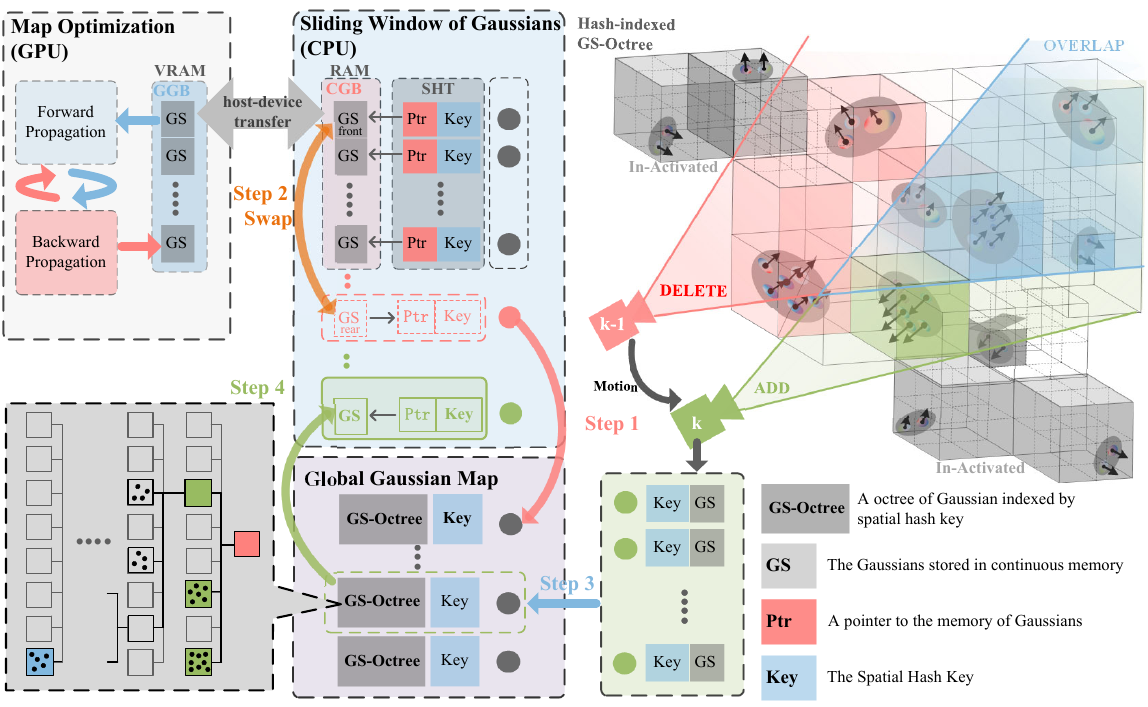}
\caption{An overview of the procedures for incrementally updating the sliding window of Gaussians (detailed in Sec.\ref{sec.map}).}
    \label{fig:iGM}
\end{figure*}

\subsection{Maintenance of Gaussian Sliding Window\label{sec.map}}

To improve memory efficiency and optimize computational speed, we limit the optimization scope to the Gaussian within the sliding window of Gaussians.

By limiting the optimization scope to the Gaussian within the sliding window of Gaussian, we can significantly enhance the optimization speed and reduce memory consumption. This targeted approach not only streamlines computational processes but also minimizes the usage of video memory, leading to more efficient performance. Additionally, this restriction helps avoid the aliasing effects associated with tile-based depth sorting in the original implementation of 3D-GS. In the original 3D-GS, tile-based depth sorting can inadvertently cause contamination of the background by foreground points, leading to visual artifacts and inaccuracies in depth representation. By confining the optimization to the Gaussian sliding window, we mitigate these aliasing issues, ensuring a cleaner separation between foreground and background elements and improving the overall quality of the depth sorting process.

Rebuilding the Gaussian sliding window from the global Gaussian map for each frame naively requires extensive memory copying, which results in significant computational overhead. However, consecutive frames typically share a large portion of the scene, making much of this effort redundant. To capitalize on the substantial overlap between frames, we introduce an incremental update strategy for the Gaussian sliding window. This method significantly reduces unnecessary memory transfers, enhances real-time performance, and scales more effectively to large and complex environments.

As illustrated in Fig.~\ref{fig:iGM}, maintaining the Gaussian sliding window involves the following key components:

\begin{itemize}
    \item \textbf{Spatial Hash Table (SHT)}: A hash-based indexing structure that maps spatial coordinates to memory pointers in CPU Memory. This ensures fast lookups and efficient organization of Gaussian parameters.
    \item \textbf{CPU Gaussian Buffer (CGB)}: A contiguous memory region in CPU Memory for storing Gaussian parameters of the currently active voxels. This compact layout enables swift data transfers to the GPU.
    \item \textbf{GPU Gaussian Buffer (GGB)}: An allocated memory block on the GPU that facilitates parallel processing and fast rendering by providing direct access to Gaussian data.
\end{itemize}

The incremental maintenance of the Gaussian sliding window involves a five-step process:

\noindent\textbf{Step 1 (Update to Global Map):}
Identify Gaussian voxels from the previous frame's sliding window of Gaussians that remain within the current FoV (\textit{OVERLAP}). Voxels that fall outside the FoV will have their optimized parameters copied back to the global Gaussian map for persistent storage, and they will be marked as \textit{DELETE}.

\noindent\textbf{Step 2 (Deletion and Compaction):}  
Swap the Gaussian parameters of voxels marked as \textit{DELETE} with those at the rear of the sliding window sequence. After repositioning all deletable voxels, remove them from the rear, thus preserving memory continuity.

\noindent\textbf{Step 3 (Overlap and Addition):}  
Using the spatial hash keys derived from the current LiDAR frame, we identify overlapping voxels (\textit{OVERLAP}) with those from the sliding window of previous frame and determine new areas that need to be integrated into the sliding window of current the FoV (\textit{ADD}). 

\noindent\textbf{Step 4 (Appending New Leaf Voxels):}  
Append all voxels (\textit{ADD}) to the rear of the CPU Gaussian Buffer (CGB) and update the Spatial Hash Table (SHT) accordingly. Subsequently, transfer the Gaussian data from the CGB (host memory) to GGB (device memory) directly, ensuring the Gaussians in the sliding window can be further optimized and rendered immediately by GPU.

This methodology significantly reduces redundant memory operations by incrementally updating only the visibility-changed voxels instead of reloading the entire sliding window. Additionally, limiting optimization to Gaussians within the current FoV decreases computational consumption, thereby enhancing real-time performance. Furthermore, leveraging the scalability and ample capacity of CPU Memory enables the system to handle larger and more complex environments with improved robustness and efficiency.

\subsection{State Estimation\label{sec.state}}

Building upon the characteristics of Gaussian rendering, we have redesigned the visual update pipeline of FAST-LIVO2~\cite{zheng2024fast}. Instead of warping patches from the current frame to the reference frame to compute the photometric error, we uniformly compute the photometric loss on the current frame by comparing the image rendered from the Gaussian map with the actual image. The convergence of this optimization is guaranteed by the smooth and differentiable nature of Gaussian rendering, as demonstrated in MonoGS~\cite{MonoGS}.

As shown in Fig. \ref{fig:Overview}, our odometry system tightly integrates LiDAR and image measurements using an IESKF with sequential updates, which is modified from FAST-LIVO2\cite{zheng2024fast}.

Our visual module is based on a semi-dense method, which is the same as~\cite{zheng2022fast, zheng2024fast}.
However, unlike Zheng et al. \cite{zheng2024fast}, we utilize a dense Gaussian map instead of a sparse visual map. 
Firstly, we employed the optimized Gaussian maps in current FoV to render a novel view with the LiDAR-inertial updated pose.

Building upon the characteristics of Gaussian rendering, we have redesigned the visual update pipeline of FAST-LIVO2. Instead of warping patches from the current frame to the reference frame to compute the photometric error, we uniformly compute the photometric loss on the current frame by comparing the image rendered from the Gaussian map with the actual image.

For the visual measurment model, we minimize the following photometric residual:

\begin{equation}
    {\TWC^*} = \arg\min_{\mathbf{T}(\boldsymbol{\xi})} 
    \sum_{{G^\text{3D}_i}\in \theta_{k-1}}
    \Big\| \mathbf{I}_k - \widehat{\mathbf{I}}_k\big(\TWC ; \boldsymbol{\theta}_{k-1} \big) \Big\|
\label{eq:minimization}
\end{equation}
where \({\theta}_{k-1}\) represents the set of \({G^\text{3D}_i}\) used to construct photometric errors, \(\boldsymbol{\xi}\) denotes the Lie algebra of the camera pose \({}^W\!\mathbf{T}_C\) to be optimized. By minimizing Equation \eqref{eq:minimization}, we iteratively optimize \(\boldsymbol{\xi}\) to make the rendered image \(\widehat{\mathbf{I}}_k\) from Gaussian map best match the observed image \(\mathbf{I}_k\).

\begin{figure}[ht]
    \centering
    \includegraphics[width=0.5\textwidth]{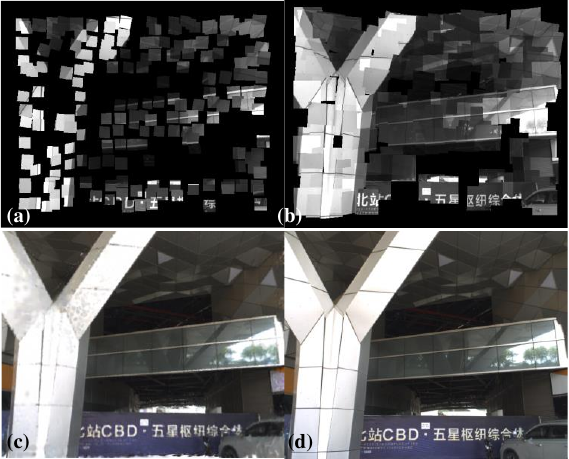}
  \caption{
  Comparison of map representation delicacy with patch based method. 
(a) and (b) illustrate the warping transformation results using patch sizes of 32 and 64, respectively. (c) presents the Gaussian rendering results, and (d) presents the reference (ground truth) image.}
    \label{fig:warped}
\end{figure}

As shown in Fig.~\ref{fig:warped}, the Gaussian map is rendered at the estimated camera pose of the current frame. For comparison, we warp the patch from the reference frame to the current frame using various patch size settings (to emulate the different levels of pyramid used in the FAST-LIVO2 method). 
With increasing patch size (and level of pyramid), it is readily observable that there appear to be seams between patches.
However, our Gaussian map representation is capable of delivering not only seamless rendering but also rendering non-Lambertian surfaces with a photorealistic quality, which highlights the advantages of our approach based on the Gaussian method.

We derive the Jacobian from photometric loss of Gaussian rendering to the IESKF-based estimator's pose update of IMU pose.

The Jacobian that relates photometric loss to camera pose is derived similarly to MonoGS~\cite{MonoGS}, as demonstrated from Equation \eqref{eqn:grad_meani_rcw1} through Equation \eqref{eqn:grad_covi_tcw1}.

\begin{align}
    \pd{\meanI}{\RotWI} = \pd{\meanI}{\meanC}\pd{\meanC}{\RotCW}
    {\pd{\RotCW}{\RotWI}}
    \label{eqn:grad_meani_rcw1}
\end{align}

\begin{equation}
    \pd{\meanI}{\tWI} = \pd{\meanI}{\meanC}\pd{\meanC}{\tCW} 
    {\pd{\tCW}{\tWI}}
    +
    \pd{\meanI}{\meanC}\pd{\meanC}{\RotCW} 
    {\pd{\RotCW}{\tWI}}
    \label{eqn:grad_meani_tcw1}
\end{equation}

\begin{align}
   \pd{\covI}{\RotWI} 
    = 
    \pd{\covI}{\matJ}\pd{\matJ}{\meanC} \pd{\meanC}{\RotCW}
    {\pd{{\RotCW}}{\RotWI}}
    +
    \pd{\covI}{\matW}
    {\pd{\matW}{\RotWI}}
\end{align}

\begin{align}
    \pd{\covI}{\tWI} 
    = \pd{\covI}{\matJ}\pd{\matJ}{\meanC} \pd{\meanC}{\tCW}
    {\pd{\tCW}{\tWI}}\notag
    +\\
    \pd{\covI}{\matJ}\pd{\matJ}{\meanC} \pd{\meanC}{\RotCW}
    {\pd{\RotCW}{\tWI}}
      \label{eqn:grad_covi_tcw1}
\end{align}

This update is systematically extended from the camera pose to the IMU pose, integrated within the IESKF framework, as demonstrated from Equation \eqref{eq:a} to Equation \eqref{eq:c}.

\begin{equation}
\pd{\RotCW}{\RotWI}
=-\RotIC^{T}
\label{eq:a}
\end{equation}

\begin{equation}
\pd{\tCW}{\tWI}=
-\RotWC^{T}
\label{eq:b}
\end{equation}

\begin{equation}
\pd{\RotCW}{\tWI}
=-{\tCI^{\wedge}}{\RotCW}
\label{eq:c}
\end{equation}

It is important to highlight that the majority of Gaussian splatting-based SLAM methods primarily depend on optimizers to compute the camera pose updates. Typically, these approaches do not assess the covariance of the updated pose; however, the pose and its covariance can be further propagated to the next sensor update such as IMU and LiDAR, and can help forming a tightly coupled IESKF system.

\section{Experiment}\label{sec.Experiment}

To provide a comprehensive assessment of the proposed system, we conducted experiments on distinct computing platforms, including a high-performance desktop and an embedded device. We first performed a comparative study against several state-of-the-art SLAM algorithms on a desktop computer (Intel i9-13900KF CPU, 128 GB RAM, and NVIDIA RTX-4090 GPU). The results demonstrate that our method achieves competitive accuracy and efficiency relative to existing approaches. Subsequently, we deployed the system on an on-board computing platform, the NVIDIA Jetson Orin NX. Despite the limited computational resources on the onboard PC, our algorithm consistently maintains real-time performance, highlighting its suitability for robotic platforms.
 
\subsection{Dataset Preparation}

In our research, we utilized multiple datasets, including public and self-collected. Among the public datasets, we selected the FAST-LIVO2 dataset~\cite{zheng2024fast}, specifically the ``CBD03'' and ``HKU01'' sequences, which depict extensive large-scale university scenes. Additionally, we employed the ``HKairport01'' and ``HKisland03'' sequences from the MARS-LVIG dataset~\cite{li2024mars}, which provide data collected in vast natural environments of mountains and seas using unmanned aerial vehicles. The MARS-LVIG dataset is distinguished by its inclusion of a D-RTK system, which provides precise ground truth for odometry.

To complement these, we collected three proprietary sequences(``Playground01,'' ``Playground02,'' and \blind{``landmark01''}) in small-scale indoor environments using a motion capture system (MoCap) as ground truth. This setup allowed us to evaluate the accuracy of our algorithm under controlled conditions. To ensure data quality, we meticulously calibrated the camera's intrinsics using a checkerboard~\cite{gao2020autonomous}, while the extrinsics between the LiDAR and camera were calibrated following the procedure in~\cite{ye2024mfcalib}. Furthermore, the temporal and spatial alignment between the optical tracker and odometry were rigorously calibrated according to the methodology in~\cite{furrer2018evaluation}. These efforts ensured that both the public and proprietary datasets were of high quality, with precise calibration and synchronization to support robust evaluation.

\subsection{Comparative Experiments}
\label{sec.Comparative}

In this section, we conduct comprehensive assessments of our system by thoroughly analyzing two critical aspects: the mapping quality through Gaussian rendering performance and the precision of the odometry.

For Gaussian-based scene reconstruction, we compare our approach with several LiDAR-assisted methods including S3Gaussian~\cite{huang2024s3gaussian} and LetsGo~\cite{LetsGo}. With respect to odometry precision, our experiments cover both traditional lateset multisensor fusion SLAM systems~\cite{lin2022r,zheng2022fast,shan2021lvi} and state-of-the-art SLAM frameworks utilizing Gaussian-based maps~\cite{keetha2024splatam,MonoGS}.

\subsubsection{Evaluation of Mapping Quality}

\begin{figure*}[ht]
    \centering
    \includegraphics[width=1.0\textwidth]{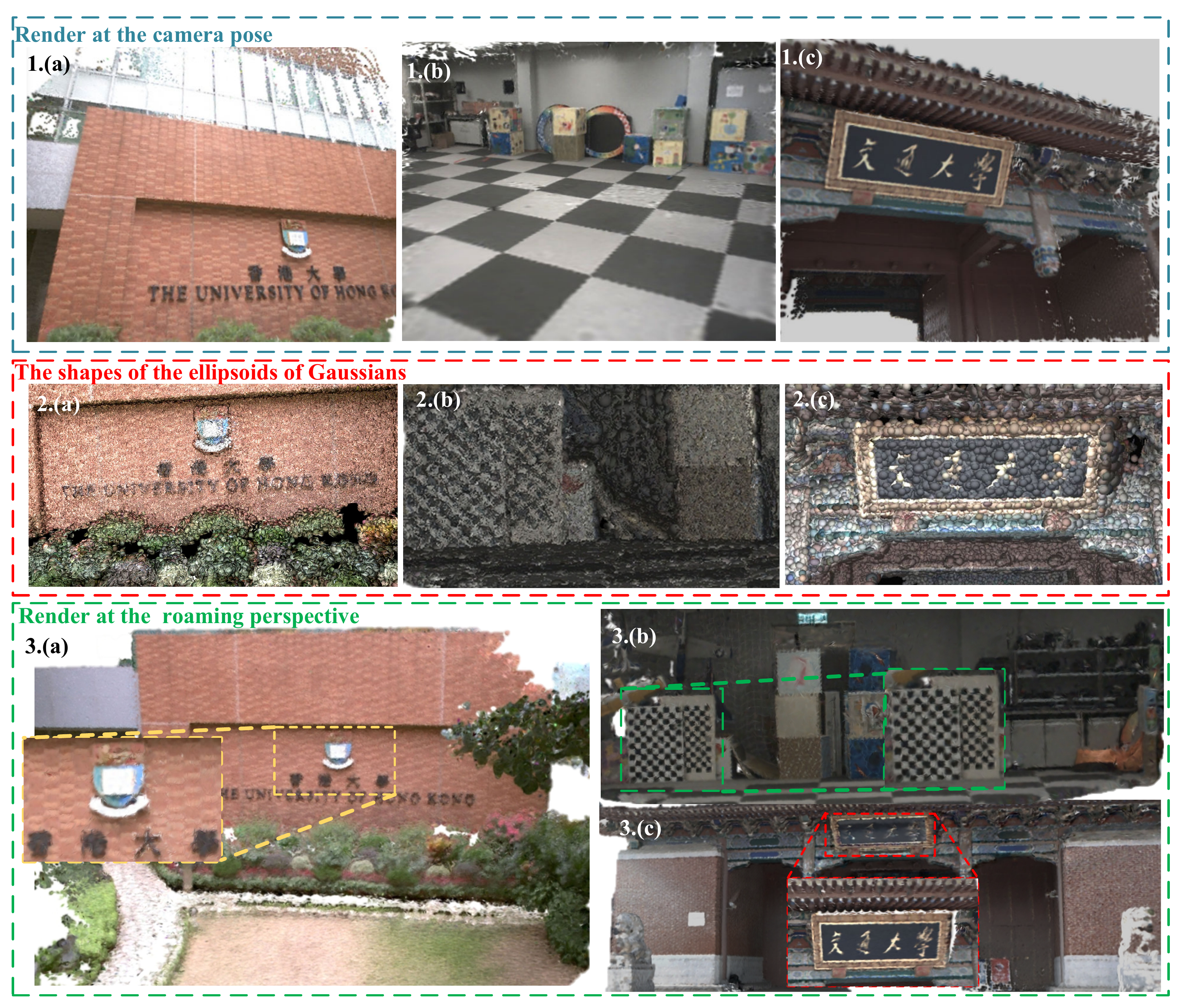}
    \caption{Mapping results of three distinct real-world scenes (a)- (c). Top row: the rendering results from camera poses. Middle row: the rendering results from roaming perspectives. Bottom row: the shapes of scene Gaussians.}
    \label{fig:T2}
\end{figure*}

For parameter settings, we employ different configurations for indoor and outdoor scenarios. In indoor environments, we use a fine root voxel size of 0.03 m with a maximum subdivision level of 2 to capture detailed features. For large-scale outdoor aerial environments, we adopt a coarser root voxel size of 1.0 m while maintaining the same subdivision level. For fair comparison, we run each method for 15,000 iterations (equivalent to 10 iterations over 1,500 frames) to ensure thorough optimization convergence.

As shown in Table~\ref{tab:render}, we first compare our approach with LiDAR-integrated Gaussian reconstruction methods such as S3Gaussian~\cite{huang2024s3gaussian} and LetsGo~\cite{LetsGo}. While LetsGo achieves adaptive LoD through distance-based voxel sizing, our fixed-level octree structure demonstrates comparable rendering quality with reduced computational overhead. The efficiency gains stem from our sliding-window strategy that enables real-time sliding-window updates, in contrast to LetsGo's offline processing approach.

Additionally, we evaluate our system against RGB-D SLAM methods utilizing Gaussian representations, specifically SplaTAM~\cite{keetha2024splatam} and MonoGS~\cite{MonoGS}. For comparison purposes, we convert LiDAR measurements to depth maps. These methods employ different strategies for map management - SplaTAM uses silhouette-based selection while MonoGS relies on covisibility-based keyframe selection. Our approach directly downsamples Gaussians in 3D voxel space, which enables more efficient extraction of structural features. Moreover, the integration of IMU measurements provides motion priors that enhance robustness against rapid movements and vibrations compared to pure RGB-D methods.

\begin{table}[h!]
\centering
\caption{Comparative Evaluation for rendering}
\label{tab:gs_comparison_datasets}
\resizebox{0.97\columnwidth}{!}{
\begin{tabular}{@{}lclccc@{}}
\toprule
Sequence        & Method      & PSNR/dB$\uparrow$ & Dur./s$\downarrow$ & Mem./GB$\downarrow$ \\ \midrule
\multicolumn{5}{c}{Indoor Datasets}                                 \\ \midrule

\textit{HKU01}~\cite{zheng2022fast}& 3D-GS~\cite{3dgstutorial}& \cellcolor{red!35}26.22    & 2128.6   & 13.8       \\
                      & SplaTAM~\cite{keetha2024splatam} & 24.06    &   292.2   & \cellcolor{orange!35}2.6       \\
                      & MonoGS~\cite{MonoGS}& 23.51   &  \cellcolor{orange!35}258.0    & 3.1       \\
                    & S3GS~\cite{huang2024s3gaussian} & \ding{55} & \ding{55}  & \ding{55} \\ 
                      & LetsGo~\cite{LetsGo}& 24.51   & 3231.3     & 18.1       \\
                      & GS-LIVO (Ours)  & \cellcolor{orange!35}25.34   & \cellcolor{red!35}82.5     & \cellcolor{red!35}2.2     \\
\midrule       
\textit{CBD03}~\cite{zheng2022fast}& 3D-GS~\cite{3dgstutorial}& \cellcolor{red!35}29.54    & 1873.8    & 12.5      \\
                      & SplaTAM~\cite{keetha2024splatam} & 26.85    & \cellcolor{orange!35}265.2     & 4.8       \\
                      & MonoGS~\cite{MonoGS}& 27.10    & 278.4     & \cellcolor{orange!35}4.6       \\ 
                    & S3GS~\cite{huang2024s3gaussian}  & 24.92 & 3450.7  & 15.2 \\ 
                      & LetsGo~\cite{LetsGo}& 25.51   & 3573.6     & 20.4       \\
                      & GS-LIVO (Ours) & \cellcolor{orange!35}27.52    & \cellcolor{red!35}88.4     & \cellcolor{red!35}2.2       \\
\midrule          
\textit{Playground01} & 3D-GS~\cite{3dgstutorial}& \cellcolor{red!35}29.75    & 763.4    & 8.8       \\
                      & SplaTAM~\cite{keetha2024splatam} & 22.01    &  292.2      & 3.4       \\
                      & MonoGS~\cite{MonoGS}& 21.15    & \cellcolor{orange!35}281.0   & \cellcolor{orange!35}3.1       \\
                      & S3GS~\cite{huang2024s3gaussian}& 20.50 & 2903.2  & 9.7 \\ 
                      & LetsGo~\cite{LetsGo}& 25.20 & 3210.3  & 10.2 \\ 
                      &GS-LIVO (Ours) &  \cellcolor{orange!35}24.09    & \cellcolor{red!35}48.5     & \cellcolor{red!35}1.5       \\
\midrule              
\textit{Playground02} & 3D-GS~\cite{3dgstutorial}&\cellcolor{red!35} 26.54    & 873.8    & 7.5      \\
                      & SplaTAM~\cite{keetha2024splatam} & 24.45    &    278.4   & \cellcolor{orange!35}3.8       \\
                      & MonoGS~\cite{MonoGS}& \cellcolor{orange!35}25.93    &  \cellcolor{orange!35}265.2 & 4.6       \\ 
                      & S3GS~\cite{huang2024s3gaussian}& 22.24  & 2957.3    & 19.1      \\
                      & LetsGo~\cite{LetsGo}& 23.12  & 2967.5    & 18.2     \\
                      & GS-LIVO (Ours) & 25.52    &\cellcolor{red!35} 63.4     & \cellcolor{red!35}1.2       \\

\midrule
\multicolumn{5}{c}{Outdoor Datasets}                                \\ \midrule

\textit{HKisland03}~\cite{li2024mars}   & 3D-GS~\cite{3dgstutorial}& \cellcolor{orange!35} 17.52    & 3494.1  & 21.6      \\
                      & SplaTAM~\cite{keetha2024splatam}& 12.60    & 790.0    & \cellcolor{orange!35}10.5      \\
                      & MonoGS~\cite{MonoGS}& 14.22    & \cellcolor{orange!35}743.7    & 12.3      \\
                      & S3GS~\cite{huang2024s3gaussian}& \ding{55} & \ding{55}  & \ding{55} \\ 
                      & LetsGo~\cite{LetsGo}&  \cellcolor{red!35}18.32 & 2803.3  & 17.6 \\ 
                      &GS-LIVO (Ours) & 15.32 & \cellcolor{red!35}82.8    & \cellcolor{red!35}3.2       \\
\midrule        
\textit{HKairport01}~\cite{li2024mars}  & 3D-GS~\cite{3dgstutorial}&\cellcolor{orange!35} 16.98    & 3919.8  & 22.8      \\
                      & SplaTAM~\cite{keetha2024splatam}& 12.39    & 915.2    & \cellcolor{orange!35}11.0      \\
                      & MonoGS~\cite{MonoGS}& 13.87    & \cellcolor{orange!35}789.6    & 13.4      \\
                      &S3GS~\cite{huang2024s3gaussian} &\ding{55}  & \ding{55}   &\ding{55}  \\ 
                      & LetsGo~\cite{LetsGo}&\cellcolor{red!35}17.32 & 3103.3  & 18.1 \\ 
                      &GS-LIVO (Ours) & 15.18    & \cellcolor{red!35}93.2    & \cellcolor{red!35}3.1       \\

\bottomrule
\end{tabular}
}
\label{tab:render}
\end{table}

Fig.~\ref{fig:T2} presents three distinct real-world scenes (HKU campus, UAV playground, and \blind{a well-known Landmark}). For each scene, we show (a) the rendering from camera view, (b) Gaussian visualization from a roaming perspective, and (c) the underlying Gaussian structure.

The first scene demonstrates high rendering fidelity with clear HKU text on the building signs. The second scene showcases precise geometry reconstruction, evidenced by sharp checkerboard patterns on the playground. In the third scene, the inscriptions are rendered with crisp detail.

The sub-figures of (c) show that the Gaussians naturally extend along surface orientations, demonstrating how our method effectively captures the scene geometry through joint LiDAR visual optimization.

Fig.~\ref{fig:T2} presents three distinct real-world scenes: (a) HKU campus, (b) UAV playground, and (c) \blind{a well-known landmark}. For each scene, we show the rendering results from camera poses (top row), results from roaming perspectives (middle row), and the underlying Gaussian structure (bottom row).
The first scene demonstrates high rendering fidelity with clear HKU text on the building signs. The second scene showcases precise geometry reconstruction, evidenced by sharp checkerboard patterns on the playground. In the third scene, the inscriptions are rendered with crisp detail.
In the bottom row for each scene, the Gaussians naturally extend along surface orientations, demonstrating how our method effectively captures the scene geometry through joint LiDAR visual optimization.

\subsubsection{Evaluation of Localization}

\begin{figure*}[ht]
    \centering
    \includegraphics[width=1.0 \textwidth]{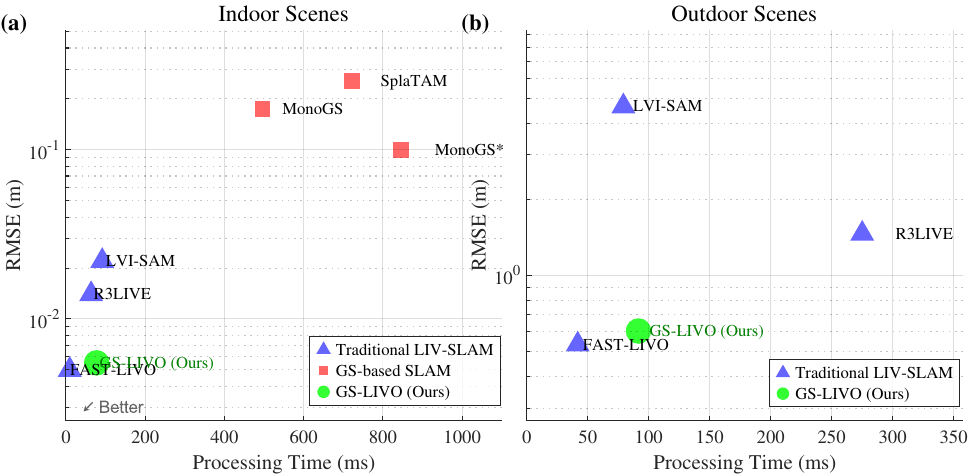}
    \caption{Performance comparison of different SLAM systems in terms of accuracy (RMSE) and computational efficiency (processing time).}
    \label{fig:star}
\end{figure*}

Fig. \ref{fig:star} presents a comprehensive performance analysis of various SLAM systems. In indoor environments (Fig. \ref{fig:star} (a)), Our method achieves comparable accuracy with traditional LIV-based SLAM methods while being significantly more efficient than other GS-based approaches. For outdoor scenarios (Fig. \ref{fig:star} (b)), our method demonstrates superior accuracy with an RMSE of 0.042m, R$^3$LIVE (1.465m), and LVI-SAM (4.665m). While the processing time of GS-LIVO is slightly higher than some traditional methods, it maintains real-time performance while providing enhanced mapping capabilities through its Gaussian Splatting representation.

For consistent evaluation across different datasets, we configure the experimental parameters as: image resolution of 640×480, octree configuration of (0.06m, 2 layers) for indoor and (0.5m, 2 layers) for outdoor environments, and a sliding window size of 100,000 Gaussians for incremental map updates.

As demonstrated in Table~\ref{tab:livslam}, GS-LIVO demonstrates competitive localization accuracy, showing notable improvements over traditional methods like R$^3$LIVE, while maintaining performance comparable to FAST-LIVO with slightly lower precision. In terms of computational efficiency, our system maintains processing times below 90 ms in both indoor and outdoor environments. This difference in processing time can be attributed to our distinct mapping approach: while FAST-LIVO~\cite{zheng2024fast} achieves efficient pose optimization through sparse visual submap warping, our system simultaneously optimizes camera poses while maintaining a photometrically accurate dense Gaussian map, requiring more computational resources for real-time dense map updates. This design choice enables us to maintain not just a hand-crafted sparse map for odometry but a photorealistic dense representation that supports high-fidelity scene reconstruction.

As presented in Table \ref{tab:gsslam}, we conducted a comprehensive comparative analysis between our proposed algorithm and state-of-the-art Gaussian-based SLAM systems, specifically SplaTAM~\cite{keetha2024splatam} and MonoGS~\cite{MonoGS}. Although MonoGS originally supports RGB-D and RGB as input, MonoGS$^{*}$~\cite{MonoGS} in our experiments denotes our implementation using LiDAR-projected depth, and MonoGS refers to its monocular version.
Unlike these Gaussian-based that rely solely on depth loss for pose estimation or a constant velocity motion model, our approach integrates LiDAR-Inertial Odometry (LIO) pose as priors, resulting in a significant improvement in accuracy. This enhances the algorithm's capability to manage the challenging motion conditions frequently faced by robotic systems.

\begin{figure}[tb]
    \centering
    \includegraphics[width=\columnwidth]{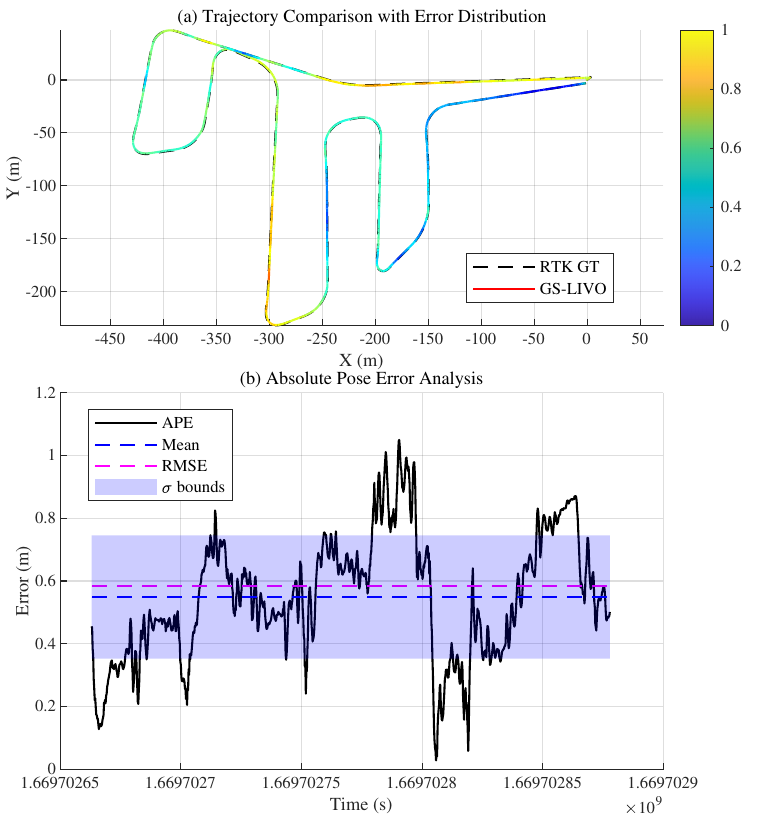}
    \caption{Trajectory and error analysis for the sequence of {HKisland03} from~\cite{li2024mars} , using RTK as ground truth. (a) The estimated trajectory color-coded by positional deviation. (b) Temporal evolution of absolute position error (APE) with mean and standard deviation bands.}
    \label{fig:path}
\end{figure}

Our system effectively handles large-scale outdoor environments, as demonstrated in Tab.~\ref{fig:path}, maintaining both real-time performance and high accuracy with a trajectory Root Mean Square Error (RMSE) of 0.58m, substantially outperforming traditional methods like R$^3$LIVE~\cite{lin2022r} and LVI-SAM~\cite{shan2021lvi}.
In particular, our computational efficiency remains competitive with traditional approaches. While R$^3$LIVE's computation time grows with map size due to increasing ikdtree indexing overhead for colored point clouds, and LVI-SAM incurs significant computational cost from its indirect method despite using a sparse map, our system maintains near real-time performance through efficient sliding window management of Gaussian points, even while maintaining a photorealistic map. In the next section, a detailed analysis of system performance is followed, including processing time and GPU memory consumption.

\begin{table}[h!]
\centering
\caption{Comparative Evaluation of LIV-based SLAM Systems Across Datasets}
\label{tab:gs_comparison_datasets}
\begin{tabular}{@{}lcccc@{}}
\toprule
Method     & Sequence       & RMSE/m$\downarrow$ & Dur./ms$\downarrow$ 
\\ \midrule
\multicolumn{4}{c}{Outdoor Datasets} \\
\midrule
FAST-LIVO~\cite{zheng2022fast}    & \textit{HKisland03 } & \cellcolor{red!35}0.51    & \cellcolor{red!35}38.9            \\
           & \textit{HKairport01  } & \cellcolor{red!35}0.56   & \cellcolor{red!35}44.5               \\
R$^3$LIVE~\cite{lin2022r}& \textit{HKisland03   } & 1.71    & 283.3             \\
           &\textit{HKairport01  } & 1.22   & 266.8             \\
LVI-SAM~\cite{shan2021lvi}& \textit{HKisland03  }  & 4.12   & \cellcolor{orange!35} 73.
5\\
           &\textit{ HKairport01  } & 5.21   &  84.8              \\
GS-LIVO (Ours) & \textit{HKisland03  } &\cellcolor{orange!35} 0.58     & 82.8           \\
           & \textit{HKairport01  } &\cellcolor{orange!35} 0.63   & \cellcolor{orange!35}93.2      \\
\midrule
\multicolumn{4}{c}{Indoor Datasets} \\
\midrule
FAST-LIVO~\cite{zheng2022fast}& \textit{Playground01 }& \cellcolor{red!35}0.005     &\cellcolor{red!35} 10.5                    \\
           & \textit{Playground02   }&\cellcolor{red!35} 0.005    &\cellcolor{red!35} 8.75                  \\
R$^3$LIVE~\cite{lin2022r}& \textit{Playground01  } & 0.014     &  60.6                 \\
           &\textit{ Playground02  } & 0.014    & 66.6                    \\
LVI-SAM~\cite{shan2021lvi}& \textit{Playground01  } & 0.023     & 96.6                \\
           & \textit{Playground02  } & 0.021    & 86.6                \\
GS-LIVO (Ours) & \textit{Playground01 }  & \cellcolor{orange!35}0.006    &\cellcolor{orange!35} 48.5                   \\
           & \textit{Playground02  } &\cellcolor{red!35} 0.005     & \cellcolor{orange!35}63.4            \\
\bottomrule
\end{tabular}
\label{tab:livslam}
\end{table}

\begin{table}[h!]
\centering
\caption{Comparative Evaluation of GS-based SLAM Systems Across Datasets}
\label{tab:gsslam}
\resizebox{0.97\columnwidth}{!}{
\begin{tabular}{@{}lcccc@{}}
\toprule
Method     & Sequence       & RMSE/m$\downarrow$ &  Dur./ms$\uparrow$ & Mem./GB$\downarrow$\\
\midrule
\multicolumn{5}{c}{Indoor Datasets}                                \\ 
\midrule
SplaTAM~\cite{keetha2024splatam}
  & \textit{Playground01} & 0.28 & 612.8 & \cellcolor{orange!35}12.5 \\
  & \textit{Playground02} & 0.23 & 831.6 & 21.0                     \\
MonoGS$^{*}$~\cite{MonoGS}
  & \textit{Playground01} & \cellcolor{orange!35}0.09 & 841.5 & 19.6  \\
  & \textit{Playground02} & \cellcolor{orange!35}0.11 & 851.2 & \cellcolor{orange!35}17.2 \\
MonoGS~\cite{MonoGS}
  & \textit{Playground01} & 0.18 & \cellcolor{orange!35}541.5 & 21.0 \\
  & \textit{Playground02} & 0.17 & \cellcolor{orange!35}451.2 & 18.7 \\
GS-LIVO (Ours) 
  & \textit{Playground01} & \cellcolor{red!35}0.006 & \cellcolor{red!35}48.5 & \cellcolor{red!35}1.2 \\
  & \textit{Playground02} & \cellcolor{red!35}0.005 & \cellcolor{red!35}63.4 & \cellcolor{red!35}1.5 \\
\midrule
\multicolumn{5}{c}{Outdoor Datasets}                                 \\ 
\midrule
SplaTAM~\cite{keetha2024splatam}
  & \textit{HKisland03} & \ding{55} & \ding{55} & \ding{55} \\
  & \textit{HKairport01} & \ding{55} & \ding{55} & \ding{55} \\
MonoGS$^{*}$~\cite{MonoGS}
  & \textit{HKisland03} & \ding{55} & \ding{55} & \ding{55} \\
  & \textit{HKairport01} & \ding{55} & \ding{55} & \ding{55} \\ 
MonoGS~\cite{MonoGS}
  & \textit{HKisland03} & \ding{55} & \ding{55} & \ding{55} \\
  & \textit{HKairport01} & \ding{55} & \ding{55} & \ding{55} \\ 
GS-LIVO (Ours)
  & \textit{HKisland03} & \cellcolor{red!35}0.58 & \cellcolor{red!35}82.8 & \cellcolor{red!35}8.0 \\
  & \textit{HKairport01} & \cellcolor{red!35}0.63 & \cellcolor{red!35}93.2 & \cellcolor{red!35}9.5 \\
\bottomrule
\end{tabular}
}
\end{table}

\begin{figure*}[ht]
    \centering
    \includegraphics[width=1.0\textwidth]{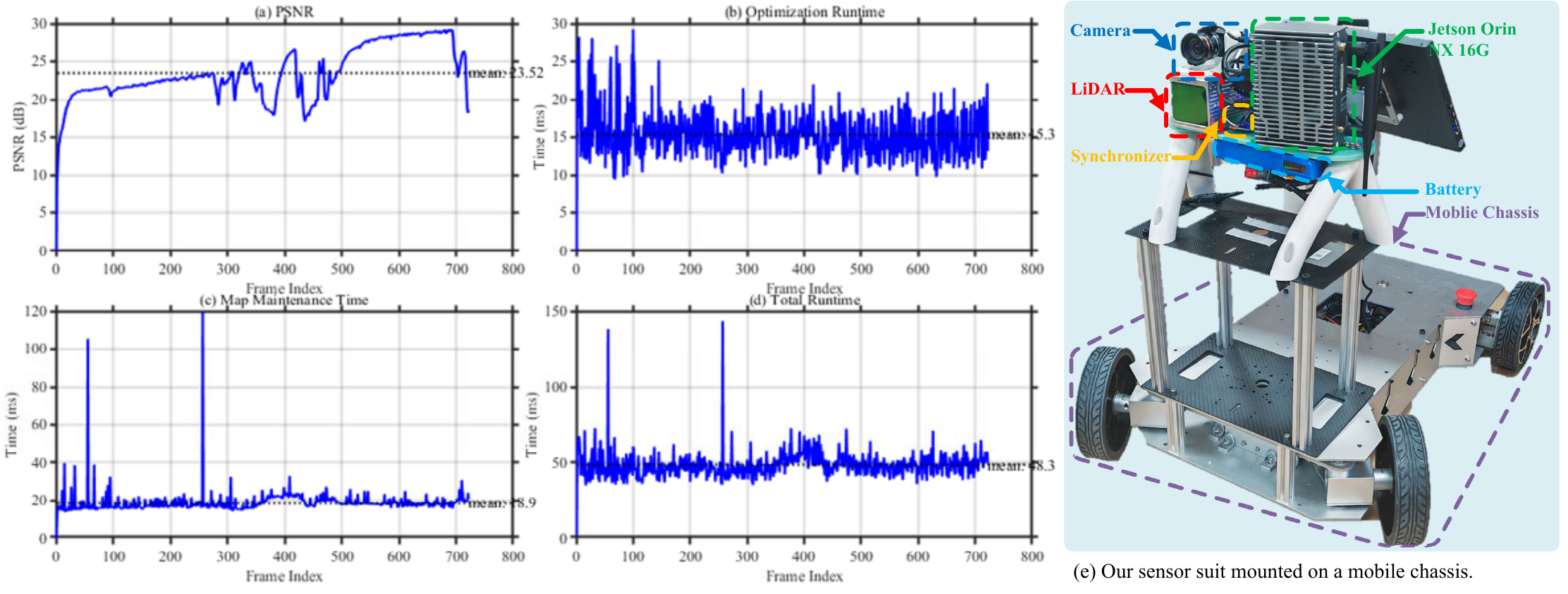}
    \caption{Performance evaluation of GS-LIVO on the embedded platform: (a)- (d) system metrics including PSNR and processing time analysis; (e)Our sensor suite integrated with Jetson Orin NX mounted on a mobile chassis.
    }
    \label{fig:Device}
\end{figure*}

\subsection{Ablation Study of Sliding Window}

\begin{figure}[ht]
    \centering
    \includegraphics[width=0.5\textwidth]{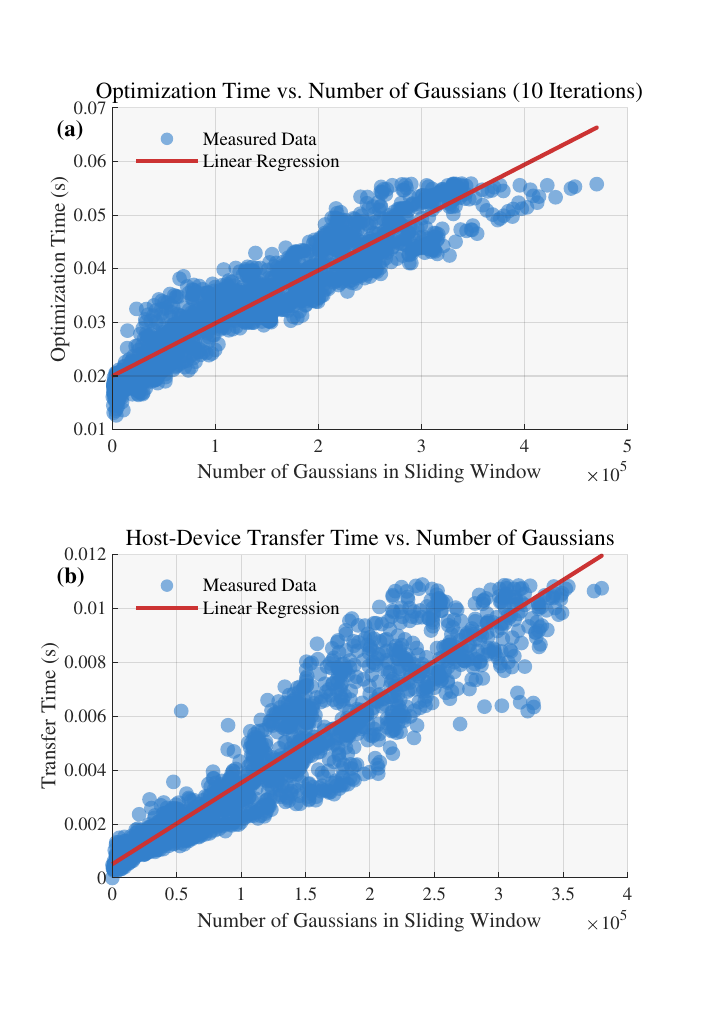}
    \caption{The relationship of the number of Gaussians with the transmission time between the GPU and CPU, as well as the optimization time (10 iterations).}
    \label{fig:time}
\end{figure}

\begin{figure}[ht]
\centering
\hspace*{-0mm}
\includegraphics[width=0.53\textwidth]{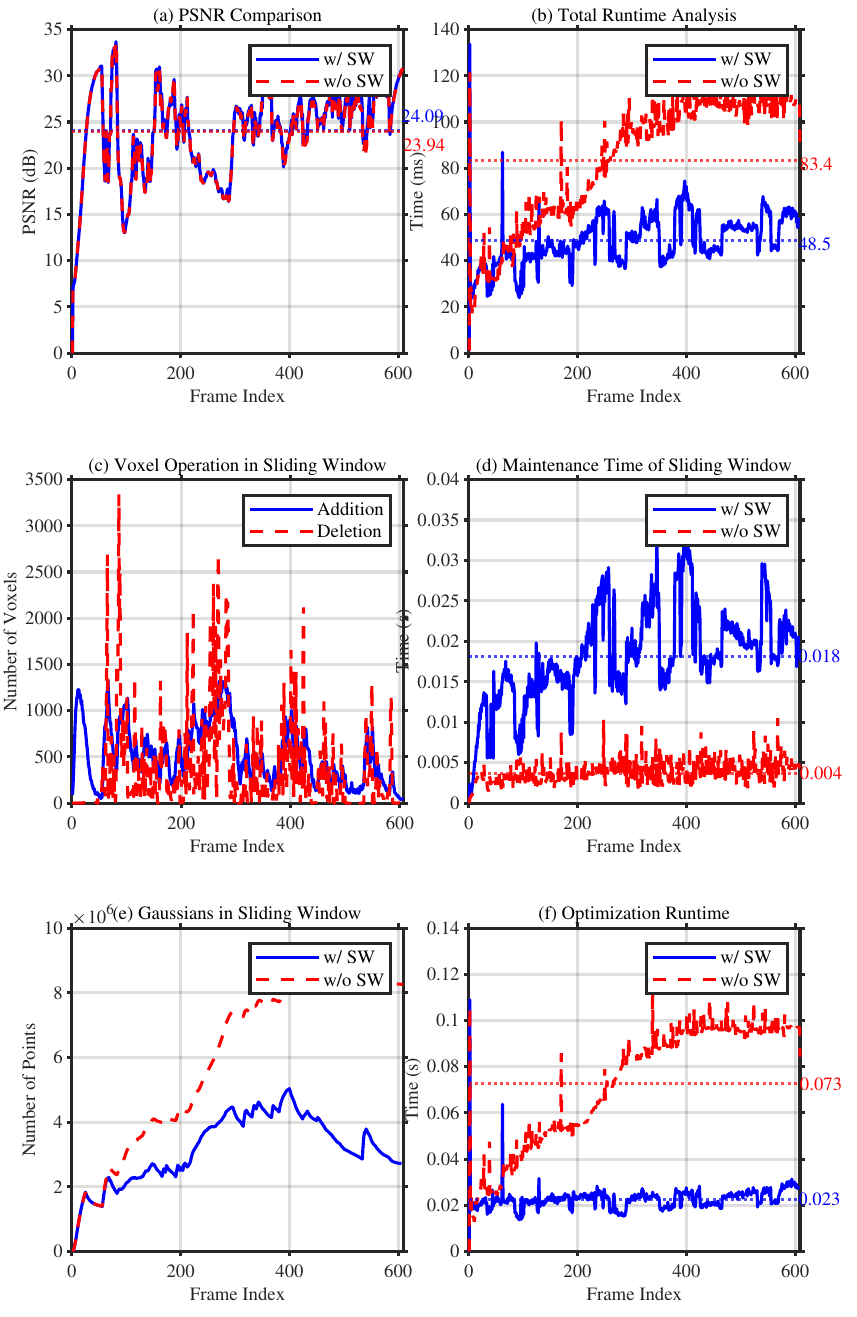}
\caption{Performance analysis of the sliding window approach in indoor environments (sequence of Playground01.bag)}
\label{fig:performance_UAV}
\end{figure}

\begin{figure}[ht] 
\centering
\hspace*{-0mm}
\includegraphics[width=0.5\textwidth]{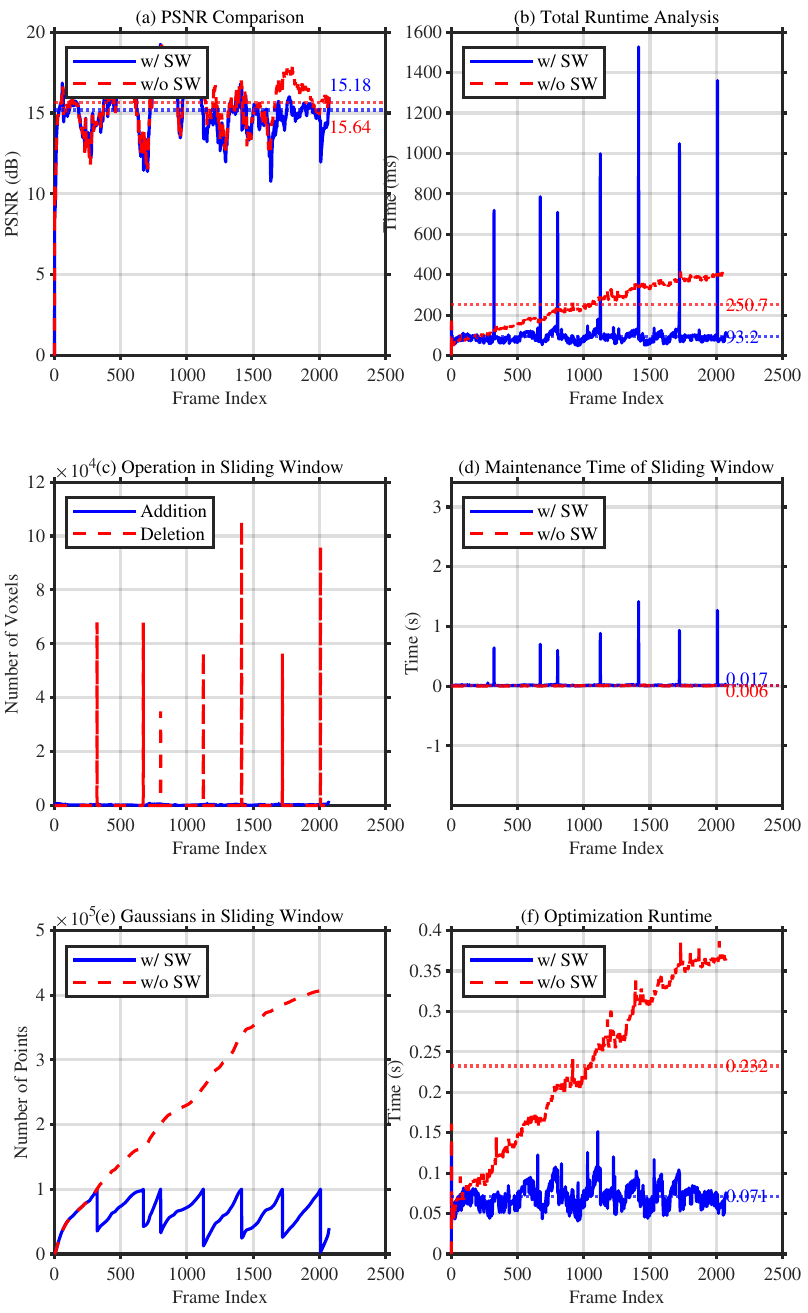}
\caption{Performance analysis of the sliding window approach in outdoor environments (sequence of HKisland03.bag)}
\label{fig:performance_M300}
\end{figure}

\label{sec.Ablation}

In our ablation study, we compared the VRAM usage and the optimization duration with and without the sliding window for Gaussian, as well as the processing time required for the map maintaining process both in indoor and outdoor sequence.

\subsubsection{Memory Consumption}

The implementation of sliding window for Gaussians shows significant advantages in both indoor (shown in Fig.~\ref{fig:performance_UAV}) and outdoor (shown in Fig.~\ref{fig:performance_M300}) environments. 
Our strategy of storing actived Gaussians of current FoV in GPU memory while maintaining the global map through an octree structure in CPU memory achieves an optimal balance between mapping quality and computational efficiency. 
As shown in Fig.~\ref{fig:performance_UAV} (a) and Fig.~\ref{fig:performance_M300} (a), this approach maintains high Peak Signal-to-Noise Ratio (PSNR) values comparable to full GPU implementations while significantly reducing memory consumption (Fig.~\ref{fig:performance_UAV} (e) and Fig.~\ref{fig:performance_M300} (e)). This efficient map management enables our system to process large-scale environments and complex scenes where traditional 3D-GS approaches would face GPU memory constraints, demonstrating the practical scalability of our method.

\subsubsection{Time Consumption}

\begin{figure}[ht]
\centering
\includegraphics[width=0.5\textwidth]{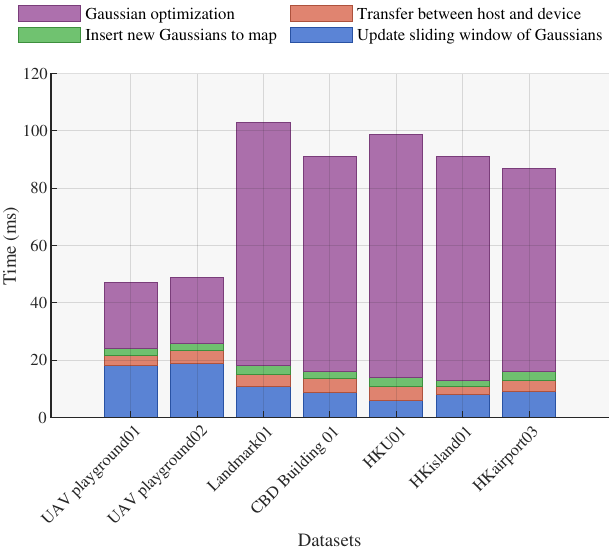}
\caption{Time consumption analysis of the mapping process}
\label{fig:run_time_analysis}
\end{figure}

As shown in Fig. \ref{fig:performance_UAV} (b) and Fig. \ref{fig:performance_M300} (b), with our sliding window strategy, the total processing time - including both window maintenance and Gaussian optimization - remains consistently below 100 ms in both indoor and outdoor environments, enabling real-time updates at 10 Hz. In contrast, approaches without sliding window optimization show steadily increasing computation times as the map grows, which hinders real-time performance in large-scale scenarios.

The breakdown of processing time, illustrated in Fig.~\ref{fig:run_time_analysis}, demonstrates that our sliding window strategy achieves efficient response times across different components. The processing overhead averages 23 ms for indoor scenes and 71 ms for outdoor environments, which can be further optimized by adjusting the Level of Detail (LoD) parameters based on available computing resources. This adaptability makes our approach suitable for various platforms and scenarios.

Importantly, as shown in Fig.~\ref{fig:performance_UAV} and Fig.~\ref{fig:performance_M300}, our sliding-window approach maintains high mapping quality while significantly reducing computational overhead. The system consistently achieves PSNR values around 25 dB, with only temporary reductions during viewpoint changes. Through iterative optimization within the sliding window, PSNR quickly recovers to values between 25 dB and 30 dB, demonstrating that our efficiency gains do not come at the cost of mapping quality.

\begin{figure}[ht]
    \centering
    \includegraphics[width=0.5\textwidth]{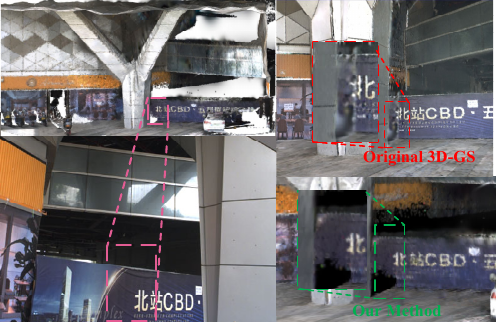}
    \caption{Comparison of aliasing artifacts in scenes with occlusions. Top: Baseline method~\cite{kerbl20233d} with highlighted aliasing region (red box). Bottom: Our method with corresponding region (green box) showing no artifacts.}
    \label{fig:alias}
\end{figure}

\subsection{Experiment on Embedded System}

To validate the efficiency of our algorithm, we deployed GS-LIVO on a mobile platform equipped with NVIDIA Jetson Orin NX (Fig.~\ref{fig:Device} (e)), configured with root voxel size of 0.5m, 2 subdivision layers, image resolution of 256×216, and a sliding window size of 20,000 Gaussians.

The system maintains real-time performance on the embedded platform (ORIN NX 16G), with optimization taking 15.3 ms (Fig.~\ref{fig:Device} (b)), map maintenance 18.9 ms (Fig.~\ref{fig:Device} (c)), and total pipeline takes 48.3 ms while achieving a PSNR of 23.52 dB (Fig.~\ref{fig:Device} (a)).

To further demonstrate the real-time and high-precision performance of our proposed odomerty, we integrated GS-LIVO into a complete autonomous navigation system. The Gaussian map is processed to generate 2D occupancy grids for path planning, while the odometry provides real-time localization for trajectory tracking. The integrated system successfully demonstrates autonomous navigation using standard planning and control algorithms (A* for global planning and LQR for trajectory tracking). Demonstration results are shown in the supplementary video.

To the best of our knowledge, this is the first real-time Gaussian-based SLAM system with online map updates deployed on an ARM-based embedded platform.

\subsection{Conclusion}

In this paper, we presented GS-LIVO, a novel real-time SLAM system that integrates traditional LiDAR-Inertial-Visual odometry with novel map representation of 3D Gaussian Splatting. By replacing conventional colored point clouds and sparse patch maps with Gaussian-based scene representation, our system achieves both accurate localization and high-fidelity mapping.
Our key contributions include: (1) a spatial hash-indexed octree structure for efficient global Gaussian map management, (2) LiDAR-visual joint initialization for high-fidelity mapping, (3) an incremental sliding-window strategy for real-time map optimization, and (4) a tightly coupled multisensor fusion framework using IESKF.

While existing Gaussian-based SLAM systems often achieve real-time localization but struggle with real-time map updates, our system leverages the advantages of traditional multi-sensor fusion to maintain high-frequency map updates while implementing tightly coupled odometry. 

Notably, GS-LIVO is the first Gaussian-based SLAM system successfully deployed on the embedded system of NVIDIA Jetson Orin NX platform, demonstrating its potential for practical robotic applications.
Extensive experiments demonstrate that GS-LIVO achieves superior performance in both indoor and outdoor environments, reducing memory consumption and optimization time while maintaining high rendering quality compared to existing methods.

Our octree-based Gaussian map effectively represents scenes; however, further research could investigate self-adaptive Level of Detail techniques that account for viewing distance, structural complexity, and texture richness. Additionally, for homogeneous regions, merging similarly colored Gaussians could further optimize memory usage and computational efficiency. However, these advanced voxel management mechanisms would introduce additional complexity that warrants careful investigation.

\bibliographystyle{support/IEEEtran}
\bibliography{root}

\begin{thebibliography}{10}
\providecommand{\url}[1]{#1}
\csname url@rmstyle\endcsname
\providecommand{\newblock}{\relax}
\providecommand{\bibinfo}[2]{#2}
\providecommand\BIBentrySTDinterwordspacing{\spaceskip=0pt\relax}
\providecommand\BIBentryALTinterwordstretchfactor{4}
\providecommand\BIBentryALTinterwordspacing{\spaceskip=\fontdimen2\font plus
\BIBentryALTinterwordstretchfactor\fontdimen3\font minus \fontdimen4\font\relax}
\providecommand\BIBforeignlanguage[2]{{%
\expandafter\ifx\csname l@#1\endcsname\relax
\typeout{** WARNING: IEEEtran.bst: No hyphenation pattern has been}%
\typeout{** loaded for the language `#1'. Using the pattern for}%
\typeout{** the default language instead.}%
\else
\language=\csname l@#1\endcsname
\fi
#2}}

\bibitem{3dgstutorial}
\BIBentryALTinterwordspacing
G.~Kopanas, B.~Kerbl, A.~Guédon, and J.~Luiten, ``{3D Gaussian Splatting Tutorial},'' 2024, international Conference on 3D Vision Tutorial. [Online]. Available: \url{https://3dgstutorial.github.io/}
\BIBentrySTDinterwordspacing

\bibitem{xu2021fast}
W.~Xu and F.~Zhang, ``Fast-lio: A fast, robust lidar-inertial odometry package by tightly-coupled iterated kalman filter,'' \emph{IEEE Robotics and Automation Letters}, vol.~6, no.~2, pp. 3317--3324, 2021.

\bibitem{xu2022fast}
W.~Xu, Y.~Cai, D.~He, J.~Lin, and F.~Zhang, ``Fast-lio2: Fast direct lidar-inertial odometry,'' \emph{IEEE Transactions on Robotics}, vol.~38, no.~4, pp. 2053--2073, 2022.

\bibitem{lin2021r}
J.~Lin, C.~Zheng, W.~Xu, and F.~Zhang, ``R2live: A robust, real-time, lidar-inertial-visual tightly-coupled state estimator and mapping,'' \emph{IEEE Robotics and Automation Letters}, vol.~6, no.~4, pp. 7469--7476, 2021.

\bibitem{lin2022r}
J.~Lin and F.~Zhang, ``R3live: A robust, real-time, rgb-colored, lidar-inertial-visual tightly-coupled state estimation and mapping package,'' in \emph{2022 International Conference on Robotics and Automation (ICRA)}.\hskip 1em plus 0.5em minus 0.4em\relax IEEE, 2022, pp. 10\,672--10\,678.

\bibitem{r3live++}
------, ``R3live++: A robust, real-time, radiance reconstruction package with a tightly-coupled lidar-inertial-visual state estimator,'' \emph{arXiv preprint arXiv:2209.03666}, 2022.

\bibitem{zheng2022fast}
C.~Zheng, Q.~Zhu, W.~Xu, X.~Liu, Q.~Guo, and F.~Zhang, ``Fast-livo: Fast and tightly-coupled sparse-direct lidar-inertial-visual odometry,'' in \emph{2022 IEEE/RSJ International Conference on Intelligent Robots and Systems (IROS)}.\hskip 1em plus 0.5em minus 0.4em\relax IEEE, 2022, pp. 4003--4009.

\bibitem{zheng2024fast}
C.~Zheng, W.~Xu, Z.~Zou, T.~Hua, C.~Yuan, D.~He, B.~Zhou, Z.~Liu, \emph{et~al.}, ``Fast-livo2: Fast, direct lidar-inertial-visual odometry,'' \emph{IEEE Transactions on Robotics}, 2024.

\bibitem{li2024mars}
H.~Li, Y.~Zou, N.~Chen, J.~Lin, X.~Liu, W.~Xu, C.~Zheng, R.~Li, D.~He, F.~Kong, \emph{et~al.}, ``Mars-lvig dataset: A multi-sensor aerial robots slam dataset for lidar-visual-inertial-gnss fusion,'' \emph{The International Journal of Robotics Research}, p. 02783649241227968, 2024.

\bibitem{Forster17troSVO}
C.~Forster, Z.~Zhang, M.~Gassner, M.~Werlberger, and D.~Scaramuzza, ``{SVO}: Semidirect visual odometry for monocular and multicamera systems,'' \emph{{IEEE} Trans. Robot.}, vol.~33, no.~2, pp. 249--265, 2017.

\bibitem{lin2023immesh}
J.~Lin, C.~Yuan, Y.~Cai, H.~Li, Y.~Ren, Y.~Zou, X.~Hong, and F.~Zhang, ``Immesh: An immediate lidar localization and meshing framework,'' \emph{IEEE Transactions on Robotics}, 2023.

\bibitem{jia2024cad}
Y.~Jia, F.~Cao, T.~Wang, Y.~Tang, S.~Shao, and L.~Liu, ``Cad-mesher: A convenient, accurate, dense mesh-based mapping module in slam for dynamic environments,'' \emph{arXiv preprint arXiv:2408.05981}, 2024.

\bibitem{zhu2024mesh}
Y.~Zhu, X.~Zheng, and J.~Zhu, ``Mesh-loam: Real-time mesh-based lidar odometry and mapping,'' \emph{IEEE Transactions on Intelligent Vehicles}, 2024.

\bibitem{ruan2023slamesh}
J.~Ruan, B.~Li, Y.~Wang, and Y.~Sun, ``Slamesh: Real-time lidar simultaneous localization and meshing,'' in \emph{2023 IEEE International Conference on Robotics and Automation (ICRA)}.\hskip 1em plus 0.5em minus 0.4em\relax IEEE, 2023, pp. 3546--3552.

\bibitem{wang2024simplified}
W.~Wang, H.~Qian, and S.~Feng, ``Simplified unstructured-mesh based uav path planning method using octree overlap detection,'' \emph{IEEE Transactions on Intelligent Vehicles}, 2024.

\bibitem{wang2020pmds}
C.~Wang, Y.~Zhang, and X.~Li, ``Pmds-slam: Probability mesh enhanced semantic slam in dynamic environments,'' in \emph{2020 5th International Conference on Control, Robotics and Cybernetics (CRC)}.\hskip 1em plus 0.5em minus 0.4em\relax IEEE, 2020, pp. 40--44.

\bibitem{cho2021sp}
H.~M. Cho, H.~Jo, and E.~Kim, ``Sp-slam: Surfel-point simultaneous localization and mapping,'' \emph{IEEE/ASME Transactions on Mechatronics}, vol.~27, no.~5, pp. 2568--2579, 2021.

\bibitem{quenzel2021real}
J.~Quenzel and S.~Behnke, ``Real-time multi-adaptive-resolution-surfel 6d lidar odometry using continuous-time trajectory optimization,'' in \emph{2021 IEEE/RSJ international conference on intelligent robots and systems (IROS)}.\hskip 1em plus 0.5em minus 0.4em\relax IEEE, 2021, pp. 5499--5506.

\bibitem{nguyen2023slict}
T.-M. Nguyen, D.~Duberg, P.~Jensfelt, S.~Yuan, and L.~Xie, ``Slict: Multi-input multi-scale surfel-based lidar-inertial continuous-time odometry and mapping,'' \emph{IEEE Robotics and Automation Letters}, vol.~8, no.~4, pp. 2102--2109, 2023.

\bibitem{wang2019real}
K.~Wang, F.~Gao, and S.~Shen, ``Real-time scalable dense surfel mapping,'' in \emph{2019 International conference on robotics and automation (ICRA)}.\hskip 1em plus 0.5em minus 0.4em\relax IEEE, 2019, pp. 6919--6925.

\bibitem{gao2023surfelnerf}
Y.~Gao, Y.-P. Cao, and Y.~Shan, ``Surfelnerf: Neural surfel radiance fields for online photorealistic reconstruction of indoor scenes,'' in \emph{Proceedings of the IEEE/CVF Conference on Computer Vision and Pattern Recognition}, 2023, pp. 108--118.

\bibitem{gao2020autonomous}
W.~Gao, K.~Wang, W.~Ding, F.~Gao, T.~Qin, and S.~Shen, ``Autonomous aerial robot using dual-fisheye cameras,'' \emph{Journal of Field Robotics}, vol.~37, no.~4, pp. 497--514, 2020.

\bibitem{qin2018vins}
T.~Qin, P.~Li, and S.~Shen, ``Vins-mono: A robust and versatile monocular visual-inertial state estimator,'' \emph{IEEE Transactions on Robotics}, vol.~34, no.~4, pp. 1004--1020, 2018.

\bibitem{mildenhall2021nerf}
B.~Mildenhall, P.~P. Srinivasan, M.~Tancik, J.~T. Barron, R.~Ramamoorthi, and R.~Ng, ``Nerf: Representing scenes as neural radiance fields for view synthesis,'' \emph{Communications of the ACM}, vol.~65, no.~1, pp. 99--106, 2021.

\bibitem{kerbl20233d}
B.~Kerbl, G.~Kopanas, T.~Leimk{\"u}hler, and G.~Drettakis, ``3d gaussian splatting for real-time radiance field rendering,'' \emph{ACM Transactions on Graphics (ToG)}, vol.~42, no.~4, pp. 1--14, 2023.

\bibitem{hierarchicalgaussians24}
\BIBentryALTinterwordspacing
B.~Kerbl, A.~Meuleman, G.~Kopanas, M.~Wimmer, A.~Lanvin, and G.~Drettakis, ``A hierarchical 3d gaussian representation for real-time rendering of very large datasets,'' \emph{ACM Transactions on Graphics}, vol.~43, no.~4, July 2024. [Online]. Available: \url{https://repo-sam.inria.fr/fungraph/hierarchical-3d-gaussians/}
\BIBentrySTDinterwordspacing

\bibitem{SucaretalICCV2021}
E.~Sucar, S.~Liu, J.~Ortiz, and A.~Davison, ``{iMAP}: Implicit mapping and positioning in real-time,'' in \emph{Proceedings of the International Conference on Computer Vision ({ICCV})}, 2021.

\bibitem{Zhu_2022_CVPR}
Z.~Zhu, S.~Peng, V.~Larsson, W.~Xu, H.~Bao, Z.~Cui, M.~R. Oswald, and M.~Pollefeys, ``Nice-slam: Neural implicit scalable encoding for slam,'' in \emph{Proceedings of the IEEE/CVF Conference on Computer Vision and Pattern Recognition (CVPR)}, June 2022, pp. 12\,786--12\,796.

\bibitem{Wang_2023_CVPR}
H.~Wang, J.~Wang, and L.~Agapito, ``Co-slam: Joint coordinate and sparse parametric encodings for neural real-time slam,'' in \emph{Proceedings of the IEEE/CVF Conference on Computer Vision and Pattern Recognition (CVPR)}, June 2023, pp. 13\,293--13\,302.

\bibitem{Johari_2023_CVPR}
M.~M. Johari, C.~Carta, and F.~Fleuret, ``Eslam: Efficient dense slam system based on hybrid representation of signed distance fields,'' in \emph{Proceedings of the IEEE/CVF Conference on Computer Vision and Pattern Recognition (CVPR)}, June 2023, pp. 17\,408--17\,419.

\bibitem{Sandstrom2023UncLeSLAM}
E.~Sandström, K.~Ta, L.~V. Gool, and M.~R. Oswald, ``Uncle-slam: Uncertainty learning for dense neural slam,'' \emph{arXiv preprint arXiv:2306.11048}, 2023.

\bibitem{Chung2022OrbeezSLAM}
C.-M. Chung, Y.-C. Tseng, Y.-C. Hsu, X.-Q. Shi, Y.-H. Hua, J.-F. Yeh, W.-C. Chen, Y.-T. Chen, and W.~H. Hsu, ``Orbeez-slam: A real-time monocular visual slam with orb features and nerf-realized mapping,'' \emph{arXiv preprint arXiv:2209.13274}, 2022.

\bibitem{keetha2024splatam}
N.~Keetha, J.~Karhade, K.~M. Jatavallabhula, G.~Yang, S.~Scherer, D.~Ramanan, and J.~Luiten, ``Splatam: Splat track \& map 3d gaussians for dense rgb-d slam,'' in \emph{Proceedings of the IEEE/CVF Conference on Computer Vision and Pattern Recognition}, 2024, pp. 21\,357--21\,366.

\bibitem{yan2023gs}
C.~Yan, D.~Qu, D.~Xu, B.~Zhao, Z.~Wang, D.~Wang, and X.~Li, ``Gs-slam: Dense visual slam with 3d gaussian splatting,'' in \emph{CVPR}, 2024.

\bibitem{sun2024mm3dgs}
L.~C. Sun, N.~P. Bhatt, J.~C. Liu, Z.~Fan, Z.~Wang, T.~E. Humphreys, and U.~Topcu, ``Mm3dgs slam: Multi-modal 3d gaussian splatting for slam using vision, depth, and inertial measurements,'' 2024.

\bibitem{xiao2024liv}
R.~Xiao, W.~Liu, Y.~Chen, and L.~Hu, ``Liv-gs: Lidar-vision integration for 3d gaussian splatting slam in outdoor environments,'' \emph{IEEE Robotics and Automation Letters}, 2024.

\bibitem{hong2024liv}
S.~Hong, J.~He, X.~Zheng, C.~Zheng, and S.~Shen, ``Liv-gaussmap: Lidar-inertial-visual fusion for real-time 3d radiance field map rendering,'' \emph{arXiv preprint arXiv:2401.14857}, 2024.

\bibitem{mueller2022instant}
\BIBentryALTinterwordspacing
T.~M\"uller, A.~Evans, C.~Schied, and A.~Keller, ``Instant neural graphics primitives with a multiresolution hash encoding,'' \emph{ACM Trans. Graph.}, vol.~41, no.~4, pp. 102:1--102:15, July 2022. [Online]. Available: \url{https://doi.org/10.1145/3528223.3530127}
\BIBentrySTDinterwordspacing

\bibitem{ORBSLAM3_TRO}
C.~Campos, R.~Elvira, J.~J. Gomez, J.~M.~M. Montiel, and J.~D. Tardos, ``{ORB-SLAM3}: An accurate open-source library for visual, visual-inertial and multi-map {SLAM},'' \emph{IEEE Transactions on Robotics}, vol.~37, no.~6, pp. 1874--1890, 2021.

\bibitem{h2mapping2023}
C.~Jiang, H.~Zhang, P.~Liu, Z.~Yu, H.~Cheng, B.~Zhou, and S.~Shen, ``H$_{2}$-mapping: Real-time dense mapping using hierarchical hybrid representation,'' \emph{IEEE Robotics and Automation Letters}, vol.~8, no.~10, pp. 6787--6794, 2023.

\bibitem{h3mapping2023}
C.~Jiang, Y.~Luo, B.~Zhou, and S.~Shen, ``H3-mapping: Quasi-heterogeneous feature grids for real-time dense mapping using hierarchical hybrid representation,'' \emph{arXiv preprint arXiv:2403.10821}, 2024.

\bibitem{swiftmapping2023}
K.~Wu, K.~Zhang, M.~Gao, J.~Zhao, Z.~Gan, and W.~Ding, ``Swift-mapping: Online neural implicit dense mapping in urban scenes,'' in \emph{Proceedings of the AAAI Conference on Artificial Intelligence}, vol.~38, no.~6, 2024, pp. 6048--6056.

\bibitem{guo2024motiongs}
X.~Guo, P.~Han, W.~Zhang, and H.~Chen, ``Motiongs: Compact gaussian splatting slam by motion filter,'' \emph{arXiv preprint arXiv:2405.11129}, 2024.

\bibitem{peng2024rtgslam}
Z.~Peng, T.~Shao, L.~Yong, J.~Zhou, Y.~Yang, J.~Wang, and K.~Zhou, ``Rtg-slam: Real-time 3d reconstruction at scale using gaussian splatting,'' 2024.

\bibitem{ren2024octree}
K.~Ren, L.~Jiang, T.~Lu, M.~Yu, L.~Xu, Z.~Ni, and B.~Dai, ``Octree-gs: Towards consistent real-time rendering with lod-structured 3d gaussians,'' \emph{arXiv preprint arXiv:2403.17898}, 2024.

\bibitem{shuai2024LoG}
Q.~Shuai, H.~Guo, Z.~Xu, H.~Lin, S.~Peng, H.~Bao, and X.~Zhou, ``Real-time view synthesis for large scenes with millions of square meters,'' 2024.

\bibitem{lang2024gaussian}
X.~Lang, L.~Li, H.~Zhang, F.~Xiong, M.~Xu, Y.~Liu, X.~Zuo, and J.~Lv, ``Gaussian-lic: Photo-realistic lidar-inertial-camera slam with 3d gaussian splatting,'' \emph{arXiv preprint arXiv:2404.06926}, 2024.

\bibitem{LetsGo}
\BIBentryALTinterwordspacing
J.~Cui, J.~Cao, F.~Zhao, Z.~He, Y.~Chen, Y.~Zhong, L.~Xu, Y.~Shi, Y.~Zhang, and J.~Yu, ``Letsgo: Large-scale garage modeling and rendering via lidar-assisted gaussian primitives,'' \emph{ACM Trans. Graph.}, vol.~43, no.~6, Nov. 2024. [Online]. Available: \url{https://doi.org/10.1145/3687762}
\BIBentrySTDinterwordspacing

\bibitem{jiang2024li}
C.~Jiang, R.~Gao, K.~Shao, Y.~Wang, R.~Xiong, and Y.~Zhang, ``Li-gs: Gaussian splatting with lidar incorporated for accurate large-scale reconstruction,'' \emph{arXiv preprint arXiv:2409.12899}, 2024.

\bibitem{hhuang2024photoslam}
H.~Huang, L.~Li, C.~Hui, and S.-K. Yeung, ``Photo-slam: Real-time simultaneous localization and photorealistic mapping for monocular, stereo, and rgb-d cameras,'' in \emph{Proceedings of the IEEE/CVF Conference on Computer Vision and Pattern Recognition}, 2024.

\bibitem{MonoGS}
H.~Matsuki, R.~Murai, P.~H.~J. Kelly, and A.~J. Davison, ``{G}aussian {S}platting {SLAM},'' in \emph{Proceedings of the IEEE/CVF Conference on Computer Vision and Pattern Recognition}, 2024.

\bibitem{zuo2019lic}
X.~Zuo, P.~Geneva, W.~Lee, Y.~Liu, and G.~Huang, ``Lic-fusion: Lidar-inertial-camera odometry,'' in \emph{2019 IEEE/RSJ International Conference on Intelligent Robots and Systems (IROS)}.\hskip 1em plus 0.5em minus 0.4em\relax IEEE, 2019, pp. 5848--5854.

\bibitem{zhu2021camvox}
Y.~Zhu, C.~Zheng, C.~Yuan, X.~Huang, and X.~Hong, ``Camvox: A low-cost and accurate lidar-assisted visual slam system,'' in \emph{2021 IEEE International Conference on Robotics and Automation (ICRA)}.\hskip 1em plus 0.5em minus 0.4em\relax IEEE, 2021, pp. 5049--5055.

\bibitem{shan2021lvi}
T.~Shan, B.~Englot, C.~Ratti, and D.~Rus, ``Lvi-sam: Tightly-coupled lidar-visual-inertial odometry via smoothing and mapping,'' in \emph{2021 IEEE international conference on robotics and automation (ICRA)}.\hskip 1em plus 0.5em minus 0.4em\relax IEEE, 2021, pp. 5692--5698.

\bibitem{hong2023rollvox}
S.~Hong, C.~Zheng, H.~Yin, and S.~Shen, ``Rollvox: Real-time and high-quality lidar colorization with rolling shutter camera,'' in \emph{2023 IEEE/RSJ International Conference on Intelligent Robots and Systems (IROS)}.\hskip 1em plus 0.5em minus 0.4em\relax IEEE, 2023, pp. 7195--7201.

\bibitem{yuan2022efficient}
C.~Yuan, W.~Xu, X.~Liu, X.~Hong, and F.~Zhang, ``Efficient and probabilistic adaptive voxel mapping for accurate online lidar odometry,'' \emph{IEEE Robotics and Automation Letters}, vol.~7, no.~3, pp. 8518--8525, 2022.

\bibitem{zwicker2001surface}
M.~Zwicker, H.~Pfister, J.~Van~Baar, and M.~Gross, ``Surface splatting,'' in \emph{Proceedings of the 28th annual conference on Computer graphics and interactive techniques}, 2001, pp. 371--378.

\bibitem{ye2024mfcalib}
T.~Ye, W.~Xu, C.~Zheng, and Y.~Cui, ``Mfcalib: Single-shot and automatic extrinsic calibration for lidar and camera in targetless environments based on multi-feature edge,'' \emph{arXiv preprint arXiv:2409.00992}, 2024.

\bibitem{furrer2018evaluation}
F.~Furrer, M.~Fehr, T.~Novkovic, H.~Sommer, I.~Gilitschenski, and R.~Siegwart, ``Evaluation of combined time-offset estimation and hand-eye calibration on robotic datasets,'' in \emph{Field and Service Robotics: Results of the 11th International Conference}.\hskip 1em plus 0.5em minus 0.4em\relax Springer, 2018, pp. 145--159.

\bibitem{huang2024s3gaussian}
N.~Huang, X.~Wei, W.~Zheng, P.~An, M.~Lu, W.~Zhan, M.~Tomizuka, K.~Keutzer, and S.~Zhang, ``S3gaussian: Self-supervised street gaussians for autonomous driving,'' \emph{arXiv preprint arXiv:2405.20323}, 2024.

\end{thebibliography}

\clearpage

\appendix
\section{Detailed Derivations for IESKF-based Photometric VIO}
\label{appendix:derivations}
This appendix presents the complete derivation chain from photometric error to IMU pose estimation in our IESKF-based system. We first establish the relationship between photometric measurements and IMU state, then detail the IESKF update framework.
\subsection{Photometric Error to IMU State}
To minimize the photometric loss with respect to the IMU pose, we need to calculate $\pd{L}{\TWI}$. As noted in~\cite{MonoGS}, the relationship between photometric error and camera pose ($\pd{L}{\TWC}$) has been established. Additionally, ~\cite{kerbl20233d} provides derivations for $\pd{L}{\covI}$ and $\pd{L}{\meanI}$, while \cite{MonoGS} gives $\pd{\covI}{\TWC}$ and $\pd{\meanI}{\TWC}$. Our work unifies these components within the IESKF framework, following approaches similar to\cite{zheng2024fast,lin2022r}.
\subsubsection{Mean Value Jacobians}
For the mean value component:
\begin{align}
\pd{\meanI}{\RotWI} &= \pd{\meanI}{\meanC}\pd{\meanC}{\RotCW}
{\pd{\RotCW}{\RotWI}}
\label{eqn:grad_meani_rcw}
\end{align}
and
\begin{equation}
\pd{\meanI}{\tWI} = \pd{\meanI}{\meanC}\pd{\meanC}{\tCW}
{\pd{\tCW}{\tWI}}+
\pd{\meanI}{\meanC}\pd{\meanC}{\RotCW}
{\pd{\RotCW}{\tWI}}
\label{eqn:grad_meani_tcw}
\end{equation}
\subsubsection{Covariance Jacobians}
For the 2D covariance terms:
\begin{align}
\pd{\covI}{\RotWI}
=
\pd{\covI}{\matJ}\pd{\matJ}{\meanC} \pd{\meanC}{\RotCW}
{\pd{{\RotCW}}{\RotWI}}
\notag\
+\\
\pd{\covI}{\matW}
{\pd{\matW}{\RotWI}}
\label{eqn:grad_covi_rcw}
\end{align}
and
\begin{align}
\pd{\covI}{\tWI}
= \pd{\covI}{\matJ}\pd{\matJ}{\meanC} \pd{\meanC}{\tCW}
{\pd{\tCW}{\tWI}}\notag\
\\
+\pd{\covI}{\matJ}\pd{\matJ}{\meanC} \pd{\meanC}{\RotCW}
{\pd{\RotCW}{\tWI}}
\label{eqn:grad_covi_tcw}
\end{align}
\subsection{Camera-IMU Transform Jacobians}
\label{subsec:camera_imu_jacobians}
To complete the above chains, we need to derive $\pd{\RotCW}{\RotWI}$, $\pd{\tCW}{\tWI}$, and $\pd{\RotCW}{\tWI}$ using the perturbation method.
\subsubsection{Perturbation Model}
The fundamental relationship between camera and IMU poses under perturbation is:
\begin{equation}
(\text{T}(\rotlll,\tlll) \cdot \TCW)^{-1} \cdot \TCI=\TWI \boxplus \text{T}{(\rotrrr,\trrr)}
\label{eq:perturbation}
\end{equation}
where $(\rotlll,\tlll)$ represents perturbations on camera pose $\TCW$ and $\xi=(\rotrrr,\trrr)$ represents perturbations on IMU pose $\TWI$.
This equation can be decomposed into rotation and translation parts:
\begin{align}
{(\Exp(\rotlll^\wedge){\RotCW})}^{-1} {\RotCI}&={\RotWI} \cdot \Exp({\rotrrr}^\wedge) \notag\
\\
\RotCW^T(
\Exp (-\rotlll^{\wedge})( \tCI-\tlll )
\tCW
)
&=
(\tWI  + \trrr)
\label{eq:decomposed}
\end{align}
\subsubsection{Rotation Component}
Starting from the rotation equation in \eqref{eq:decomposed}:

Taking inverse and substituting:
\begin{equation}
{\RotWC} \Exp(-{\rotlll}^\wedge) {\RotCI} = {\RotWI} \Exp({\rotrrr}^\wedge)
\end{equation}
Using first-order approximation $\Exp(\xi^\wedge) \approx I + \xi^\wedge$:
\begin{equation}
\RotWI - \RotWC \rotlll^\wedge \RotCI = \RotWI + \RotWI \rotrrr^\wedge
\end{equation}
After cancellation:
\begin{align}
-\RotIC \rotlll^\wedge \RotCI &= \rotrrr^\wedge \\
(-\RotIC \rotlll)^\wedge &= \rotrrr^\wedge \\
\rotlll &= -\RotCI \rotrrr
\end{align}

Therefore:
\begin{equation}
\pd{\RotCW}{\RotWI} = -\RotCI
\label{eq:rot_jacobian}
\end{equation}
\subsubsection{Translation Component}
Starting from the translation equation in \eqref{eq:decomposed}:

Expanding first-order terms:
\begin{align}
{\RotWC}(I-\rotlll^{\wedge})\tCI
-{\RotWC}(I-\rotlll^{\wedge})(\tCW + \tlll)
= \tWI + \trrr
\label{eq:trans_expand}
\end{align}
After algebraic simplification:
\begin{equation}
{\RotWC}\tCI^{\wedge}\rotlll - {\RotWC}\tlll = \trrr
\label{eq:trans_simple}
\end{equation}

Therefore:
\begin{align}
\pd{\tCW}{\tWI} &= -\RotCW \label{eq:trans_jacobian1} \\
\pd{\RotCW}{\tWI} &= -\tCI^{\wedge}\RotCW \label{eq:trans_jacobian2}
\end{align}
\subsection{IESKF Update Framework}
\subsubsection{Prior and Posterior State}
The state consists of:
\begin{itemize}
\item Prior state: $\bar{\TWI}$ with covariance $\boldsymbol{\Sigma}_{\bar{\TWI}}$
\item Posterior state: $\check{\TWI}$ with covariance $\boldsymbol{\Sigma}_{\check{\TWI}}$
\end{itemize}
The Jacobians derived in Section~\ref{subsec:camera_imu_jacobians} relate to $\mathbf{J}_i$ in the measurement model:
\begin{equation}
\mathbf{M}(\hat{\TWI}, \textbf{u}_{i}) + \mathbf{J}_i \xi
\end{equation}
\subsubsection{Update Process}
The IESKF update optimizes:
\begin{align}
\arg\min_{\delta\hat{\TWI}} &\sum{\mathbf{u}_i} \Big\|\mathbf{M}(\hat{\TWI}, \mathbf{u}_{i}) + \mathbf{J}_i  T(\xi) \Big\|^2_{\boldsymbol{\Sigma}{{u}_i}}
\notag \\
&+ \Big\| \hat{\TWI}\boxminus \bar{\TWI} + \boldsymbol{\mathcal{H}}  T(\xi)
\Big\|^2_{\boldsymbol{\Sigma}_{\bar{\TWI}}}
\label{eq:ieskf_objective}
\end{align}
The solution follows:
\begin{align}
\mathbf{H} &= [\mathbf{J}_1^T, \cdots, \mathbf{J}_m^T]^T \\
\mathbf{R} &= \text{diag}(\boldsymbol{\Sigma}{{u}_1}, \cdots, \boldsymbol{\Sigma}{{u}_m}) \\
\mathbf{P} &= \boldsymbol{\mathcal{H}}^{-1} \boldsymbol{\Sigma}_{\bar{\TWI}} \boldsymbol{\mathcal{H}}^{-T} \\
\mathbf{K} &= (\mathbf{H}^T\mathbf{R}^{-1}\mathbf{H} + \mathbf{P}^{-1})^{-1}\mathbf{H}^T\mathbf{R}^{-1}
\end{align}
The state and covariance are updated as:
\begin{align}
\check{\TWI} &= \hat{\TWI} \boxplus {T(\xi)} \\
\boldsymbol{\Sigma}_{\check{\TWI}} &= (\mathbf{I} - \mathbf{KH})\mathbf{P}
\end{align}
where:
\begin{equation}
{T(\xi)} = -\mathbf{K}\mathbf{z} - (\mathbf{I}- \mathbf{KH})(\boldsymbol{\mathcal{H}})^{-1}(\hat{\TWI}\boxminus \bar{\TWI})
\end{equation}

\end{document}